\documentclass[lettersize,journal]{IEEEtran}
\usepackage{amsmath,amsfonts}
\usepackage{algorithmic}
\usepackage{algorithm}
\usepackage{array}
\usepackage[caption=false,font=normalsize,labelfont=rm,textfont=rm]{subfig}
\usepackage{textcomp}
\usepackage{stfloats}
\usepackage{url}
\usepackage{verbatim}
\usepackage{graphicx}
\usepackage{cite}
\usepackage{multirow}
\usepackage{amsfonts,amssymb} 
\usepackage{enumerate}
\usepackage{color}
\usepackage[table,xcdraw,dvipsnames,svgnames,x11names]{xcolor}
\usepackage[normalem]{ulem}
\useunder{\uline}{\ul}{}
\usepackage{pifont}
\usepackage{booktabs}
\DeclareSymbolFont{mathbf}{OT1}{cmr}{bx}{n}
\DeclareMathSymbol{X}{\mathalpha}{mathbf}{`X}
\DeclareMathSymbol{Z}{\mathalpha}{mathbf}{`Z}
\DeclareMathSymbol{Y}{\mathalpha}{mathbf}{`Y}
\DeclareMathSymbol{W}{\mathalpha}{mathbf}{`W}
\DeclareMathSymbol{E}{\mathalpha}{mathbf}{`E}
\begin{document}

\title{JL1-CD: A New Benchmark for Remote Sensing Change Detection and a Robust Multi-Teacher Knowledge Distillation Framework}

\author{Ziyuan Liu, Ruifei Zhu, Long Gao, Yuanxiu Zhou, Jingyu Ma, and Yuantao Gu,~\IEEEmembership{Senior Member,~IEEE}
\thanks{Ziyuan Liu and Yuantao Gu are with the Department of Electronic Engineering, Beijing National Research Center for Information Science and Technology, Tsinghua University, Beijing 100084, China (e-mail: liuziyua22@mails.tsinghua.edu.cn; gyt@tsinghua.edu.cn). 
Ruifei Zhu, Long Gao, Yuanxiu Zhou, and Jingyu Ma are with Chang Guang Satellite Technology Co., Ltd. (CGSTL)
Changchun 130102, China (e-mail: zhuruifei@jl1.cn; gaolong1056@jl1.cn; zhouyuanxiu@jl1.cn; majingyu@jl1.cn). 
(\textit{Corresponding author: Yuantao Gu.})}
}



\maketitle
\begin{abstract}
Change detection (CD) in remote sensing images plays a vital role in Earth observation. However, the scarcity of high-resolution, comprehensive open-source datasets and the difficulty in achieving robust performance across varying change types remain major challenges. To address these issues, we introduce JL1-CD, a large-scale, sub-meter CD dataset consisting of 5,000 image pairs. We further propose a novel Origin-Partition (O-P) strategy and integrate it into a Multi-Teacher Knowledge Distillation (MTKD) framework to enhance CD performance. The O-P strategy partitions the training set by Change Area Ratio (CAR) and trains specialized teacher models on each subset. The MTKD framework then distills complementary knowledge from these teachers into a single student model, enabling improved detection results across diverse CAR scenarios without additional inference cost. Our MTKD approach demonstrated strong performance in the 2024 ``Jilin-1'' Cup challenge, ranking first in the preliminary and second in the final rounds. Extensive experiments on the JL1-CD and SYSU-CD datasets show that the MTKD framework consistently improves the performance of CD models with various network architectures and parameter sizes, establishing new state-of-the-art results. Code and dataset are available at \url{https://github.com/circleLZY/MTKD-CD}.
\end{abstract}

\begin{IEEEkeywords}
Benchmark, change detection, knowledge distillation, remote sensing.
\end{IEEEkeywords}

\section{Introduction} \label{Introduction}
Remote sensing image change detection (CD) is a technique used to detect and analyze surface changes by leveraging multi-temporal data \cite{AdaptFormer}. 
Over the past few decades, it has been extensively studied and has become a crucial tool for Earth surface observation. 
CD plays a significant role in various fields, including land-use change updates, natural disaster assessment, environmental monitoring, and urban planning.

In recent years, the rapid advancement of deep learning (DL) has revolutionized remote sensing CD, delivering substantial performance breakthroughs. 
Convolutional Neural Networks (CNNs), which have demonstrated remarkable success in image processing, were the first neural network architecture applied to remote sensing CD and remain widely used and optimized today \cite{fcsn, ifn, stanet, snunet, changestar, tinycd, hanet, cgnet, lightcdnet, seifnet}. 
With the introduction of Transformers, several studies have explored their application in CD tasks \cite{changeformer,changer,ctd-former, FTA-Net}. 
The Mamba architecture, a state space sequence model designed for efficient long-range dependency modeling, has attracted growing interest as a promising alternative to Transformers in CD tasks \cite{changemamba, CDMamba}.
More recently, Foundation Model (FM) has emerged as a novel paradigm, aiming to achieve multi-task and multi-domain generalization through large-scale pretraining \cite{changeclip,optimizing,ban,ttp,anychange}. 

However, DL-based CD methods generally face two major challenges: the scarcity of high-quality, high-resolution, all-inclusive CD datasets and limitations in handling highly diverse change scenarios. 
Although numerous CD datasets have been proposed, they are often tailored to specific scenarios, thus restricting the generalization capabilities of the algorithms. 
For instance, models trained on datasets focused on human-induced changes often fail to perform effectively when confronted with natural change scenarios. 
On the other hand, the learning capacity of these models is inherently limited. 
Most existing algorithms rely on a single-phase training approach, typically end-to-end training.
Although such training strategies perform well on limited change types, model accuracy degrades significantly as change diversity increases.

To address the aforementioned challenges, we construct a new large-scale, high-resolution, all-inclusive open-source CD dataset, JL1-CD, which is named after the Jilin-1 satellite.
This dataset comprises 5,000 pairs of 512 × 512 images captured in China, with a resolution of 0.5–0.75 meters, along with binary change labels at the pixel level. 
The JL1-CD dataset not only includes common human-induced changes such as buildings and roads but also encompasses various natural changes, such as forests, water bodies, and grasslands. 
To quantitatively characterize the diversity of change areas in datasets, we propose the Change Area Ratio (CAR) metric and partition the dataset into subsets based on different CAR levels, which we refer to as the Original-Partition (O-P) strategy.
By training separate CD models on each partition, the O-P strategy effectively simplifies the learning task and improves detection accuracy.
Building upon this, we further propose the Multi-Teacher Knowledge Distillation (MTKD) framework that distills complementary knowledge from these partition-specific teacher models into a single student model.
This student model combines the strengths of individual teachers, achieving superior detection performance without increasing resource consumption during inference.

\begin{figure*}[!t]
    \centering
    \includegraphics[width=0.95\textwidth]{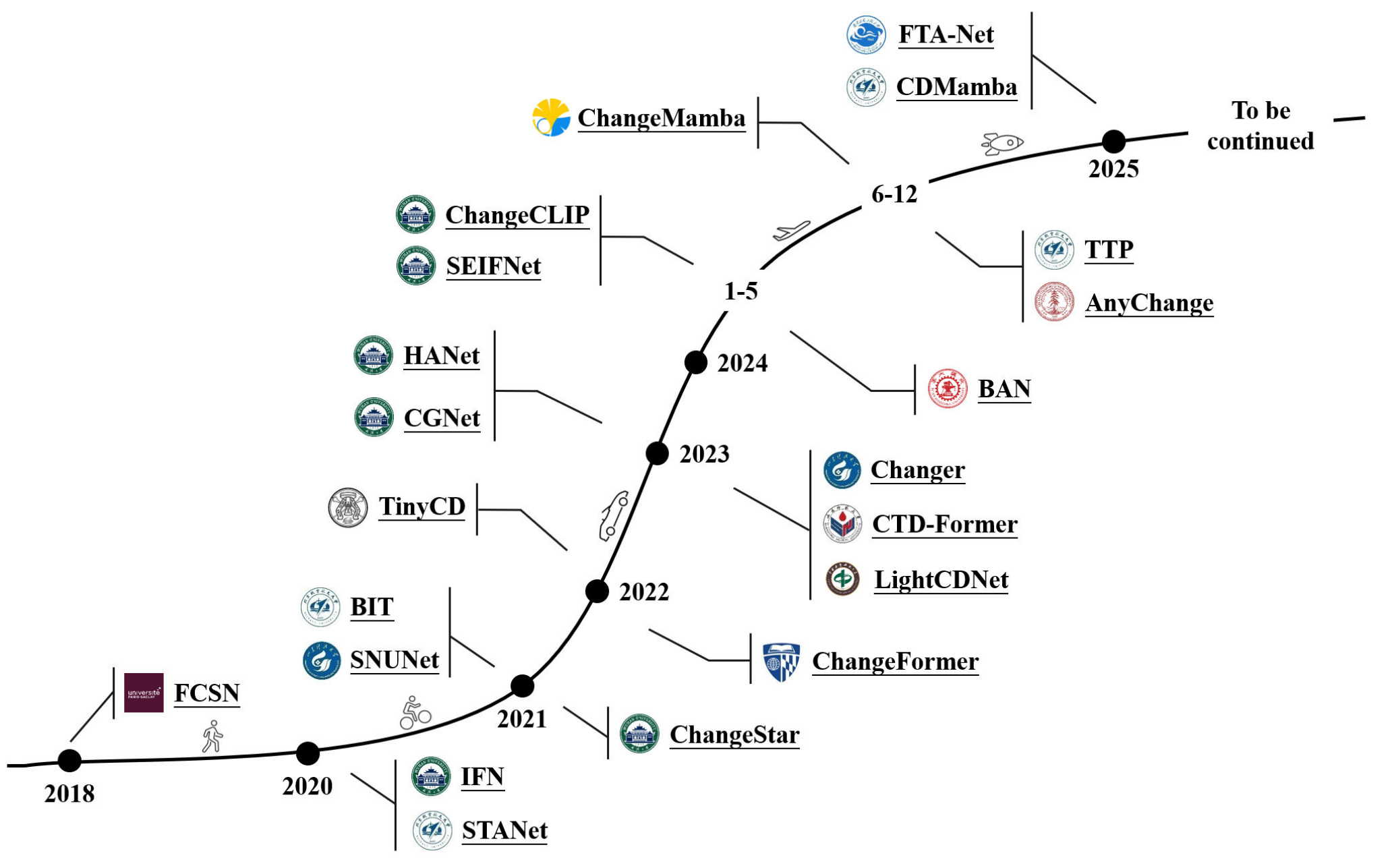}
    \caption{Timeline of the development of mainstream DL-based CD methods. 
    }
    \label{fig:survey}
\end{figure*}

Our main contributions are as follows: 
\begin{enumerate}[1)]
\item We introduce JL1-CD, a new sub-meter, all-inclusive open-source CD dataset.
\item We evaluate numerous existing algorithms on the JL1-CD dataset, establishing a new benchmark for remote sensing CD by publicly releasing the associated dataset, code and models. This contribution is anticipated to facilitate further advancements in the field.
\item We propose MTKD, a general optimization framework for CD models, which achieves consistent improvement across diverse CD models with varying architectures and parameter scales on different datasets.
\end{enumerate}

\section{Related Works} \label{Related Works}
\subsection{DL-Based CD}
Deep learning has experienced rapid advancements, achieving remarkable success in remote sensing image CD. 
As illustrated in Fig. \ref{fig:survey}, we present a timeline of the development of mainstream DL-based CD algorithms. 
Based on the differences in neural network architectures and training paradigms, we briefly review three major categories of DL-based CD methods that are widely adopted in current research.

\paragraph{CNN-Based CD}
CNNs serve as the foundation of many early DL-based CD methods and remain widely used today. 
Daudt et al. \cite{fcsn} proposed three fully convolutional neural network architectures, marking the entry of DL into CD research. 
This breakthrough spurred numerous subsequent CNN-based solutions. More recently, researchers have also developed efficient CD models like TinyCD \cite{tinycd} and LightCDNet \cite{lightcdnet} that achieve superior performance while being significantly smaller and faster.

\paragraph{Transformer-Based CD}
Transformer-based methods have emerged as a promising approach for CD. 
Chen et al. \cite{bit} introduced the bi-temporal image transformer (BIT), combining a transformer encoder with a ResNet backbone to model spatial-temporal contexts efficiently. 
Bandara et al. \cite{changeformer} proposed ChangeFormer, a fully transformer-based Siamese network for CD, which unifies a hierarchical transformer encoder with a multi-layer perceptron (MLP) decoder.
Fang et al. \cite{changer} introduced the Changer series framework, a novel architecture for CD that incorporates alternative interaction layers between bi-temporal features. 
This framework is applicable to both CNN-based and Transformer-based models, enhancing the performance of the original models.

\paragraph{FM-Based CD} 
Recently, foundation models have become a new training paradigm.
There have been works utilizing remote sensing data to fine-tune pretrained models such as Vision Transformers (ViT) \cite{vit}, Segment Anything Model (SAM) \cite{sam}, and Contrastive Language-Image Pretraining (CLIP) \cite{clip}, achieving better performance in CD tasks \cite{ban,changeclip,ttp}.
Li et al. \cite{ban} proposed the Bi-Temporal Adapter Network (BAN), a universal FM-based framework for CD, which enhances existing models with minimal additional parameters and achieves significant performance improvements.
Chen et al. \cite{ttp} introduced Time Travelling Pixels (TTP), a method that integrates latent knowledge from the SAM model into CD, overcoming domain shifts and spatio-temporal complexities.

\begin{figure*}[t]
    \centering
    \includegraphics[width=1.0\textwidth]{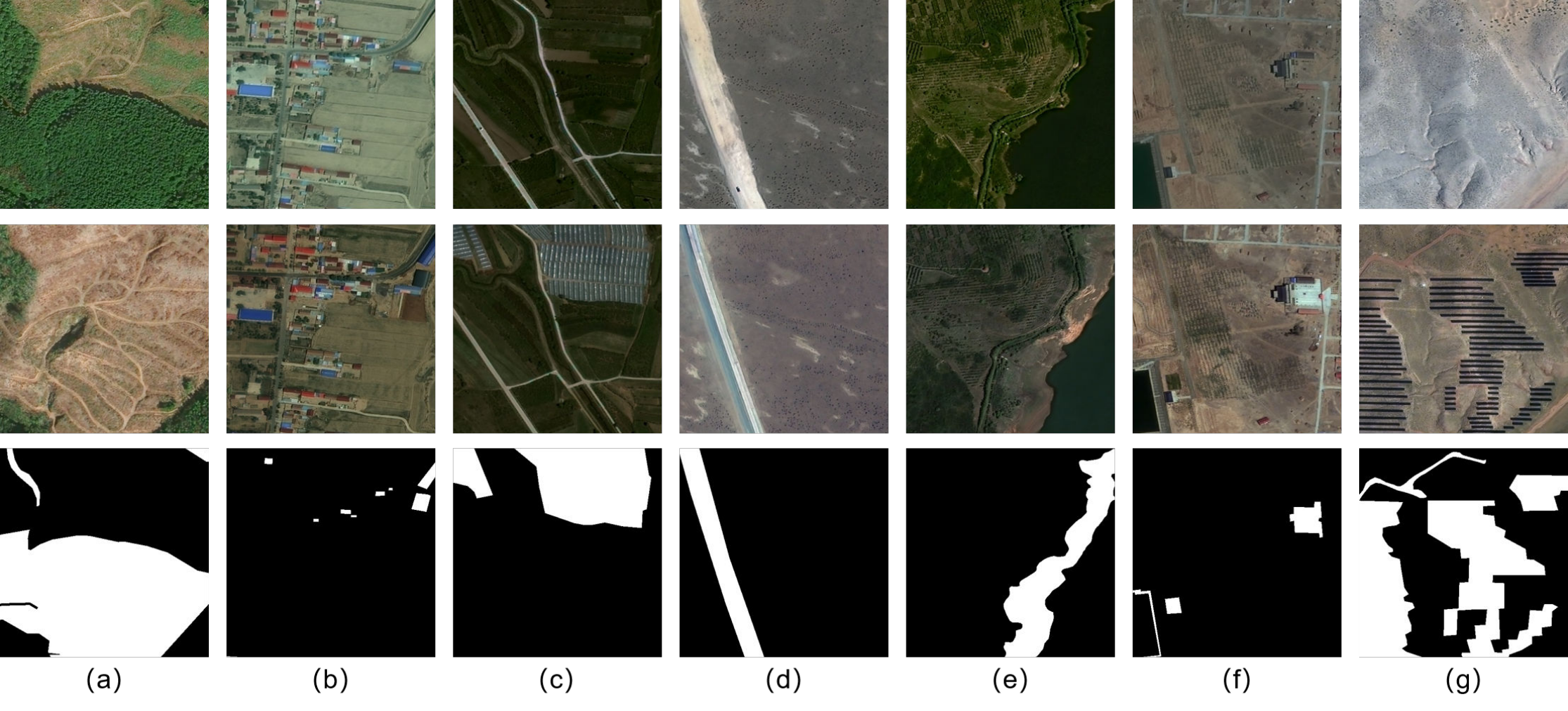}
    \caption{Representative samples from JL1-CD dataset. Each column, from top to bottom, represents the pre-event image, the post-event image, and the ground truth label. Each column corresponds to different change types: (a) decrease in woodland; (b) building changes; (c) conversion of cropland to greenhouses; (d) road changes; (e) waterbody changes; (f) surface hardening; and (g) photovoltaic panel construction.}
    \label{fig:dataset-class}
\end{figure*}

\begin{table}[!t]
\centering
\caption{Information of Open-Source Binary CD Datasets and the Proposed JL1-CD Dataset\label{table:dataset_information}}
\renewcommand{\arraystretch}{1.25}
\setlength{\tabcolsep}{5.8pt}
\begin{tabular}{lccccc}
\hline
Dataset & Year & Image Pairs & Image Size & Resolution \\ \hline
SZTAKI \cite{sztaki}      & 2009 & 13  & \begin{tabular}[c]{@{}c@{}}$952\times640$\\ $1,048\times724$\end{tabular} & 1.5 \\
CDD \cite{cdd}            & 2011 & 16,000      & 256 × 256  & 0.03-1 \\
WHU-CD \cite{whucd}       & 2018 & 1    & 32,20 × 15,354 & 0.2    \\
DSIFN \cite{ifn}          & 2020 & 394         & 512 × 512  & 2  \\
LEVIR-CD \cite{stanet}    & 2020 & 637  & 1,024 × 1,024   & 0.3   \\
S2Looking \cite{s2looking}& 2021 & 5,000 & 1,024 × 1,024 & 0.5-0.8 \\
SYSU-CD \cite{sysucd}     & 2021 & 20,000  & 256 × 256 & 0.5  \\
\rowcolor[gray]{0.92} 
JL1-CD                   & 2025 & 5,000   & 512 × 512  & 0.5-0.75     \\ \hline
\end{tabular}
\end{table}

\subsection{Knowledge Distillation in CD}
Knowledge distillation (KD), introduced by Hinton et al. \cite{hinton}, aims to transfer the representational knowledge of a teacher network to a smaller student network. 
In recent years, as the complexity of DL models in remote sensing tasks has increased, researchers have explored how to transfer knowledge from large, complex teacher models to smaller, more efficient student models through KD, thereby improving performance \cite{dsrkd, sar}.

Yan et al. \cite{distill1} proposed a novel self-supervised learning approach for unsupervised CD by fusing domain knowledge of remote sensing indices during both training and inference. 
Wang et al. \cite{distill2} addressed remote sensing semantic CD (SCD), which focuses on detecting changes in land cover and land use over time. 
Wang et al. \cite{distill3} proposed a KD-based method for CD (CDKD), designed to overcome the challenges of deploying large deep learning models with high computational and storage requirements on resource-constrained spaceborne edge devices. 
Although these methods have successfully utilized KD to enhance the performance of various student models, they are tailored to specific models and do not provide a generalized distillation framework applicable to other CD models. 
Furthermore, open-source KD-based implementations for remote sensing CD tasks remain limited.

In contrast, the proposed MTKD framework substantially improves the performance of CD models with various architectures and parameter sizes, and we commit to open-sourcing all the code and models.

\section{JL1-CD Dataset} \label{JL1-CD Dataset}
High-resolution, all-inclusive CD datasets are crucial for remote sensing applications. High-resolution images provide richer spatial information, which is more conducive to visual interpretation compared to medium- and low-resolution images. 
Datasets with comprehensive change features enable the development of algorithms with greater generalization and transferability.  
Despite the numerous open-source change detection datasets proposed over the past decades, many still lack sub-meter-level resolution, and the variety of change types remains limited. 
These limitations hinder progress in CD research, particularly in the development of DL-based algorithms.

Table \ref{table:dataset_information} summarizes mainstream binary CD datasets in terms of release year, number of image pairs, image size, and spatial resolution.
The SZTAKI AirChange Benchmark \cite{sztaki} contains 12 pairs of 952 × 640 and one pair of 1,048 × 724 optical aerial images. It is one of the earliest and most widely used CD datasets in early research. 
The DSIFN dataset \cite{ifn} consists of 6 large bi-temporal image pairs from 6 cities in China, which are cropped into 394 sub-image pairs, each sized 512 × 512. 
However, the resolution of all or part of the images in these datasets does not reach the sub-meter level.  
WHU-CD \cite{whucd}, LEVIR-CD \cite{stanet}, and S2Looking \cite{s2looking} are widely recognized datasets focusing on monitoring building changes. 
These datasets predominantly include human-induced changes and lack natural change types. 
The CDD dataset \cite{cdd} is derived from 7 pairs of 4,725 × 2,700 real-world seasonal change remote sensing images. 
The SYSU-CD dataset \cite{sysucd} contains 20,000 pairs of 0.5-meter aerial images captured in Hong Kong between 2007 and 2014 with a size of 256 × 256.
While both datasets offer high resolution and diverse change types, the image acquisition dates back several years, thus not reflecting the most recent surface changes.

To provide a new benchmark for evaluating CD algorithms, we propose the JL1-CD dataset, a high-resolution, all-inclusive change detection dataset. 
JL1-CD includes 5,000 pairs of satellite images captured in China from early 2022 to the end of 2023, including Shandong, Ningxia, Anhui, Hebei, Hunan, and other regions.
The images have sub-meter resolutions ranging from 0.5 to 0.75 meters and are sized 512 × 512 pixels. 
As shown in Fig. \ref{fig:dataset-class}, the dataset covers various common human-induced and natural surface features, such as buildings, roads, hardened surfaces, woodlands, grasslands, croplands, water bodies and photovoltaic (PV) panels.
The original 5,000 image pairs are divided into 4,000 pairs for training and 1,000 pairs for testing, following a split of 80\%: 20\%.  
The JL1-CD dataset has been made openly available for all research needs.

\begin{figure*}[!t]
    \centering
    \includegraphics[width=1.0\textwidth]{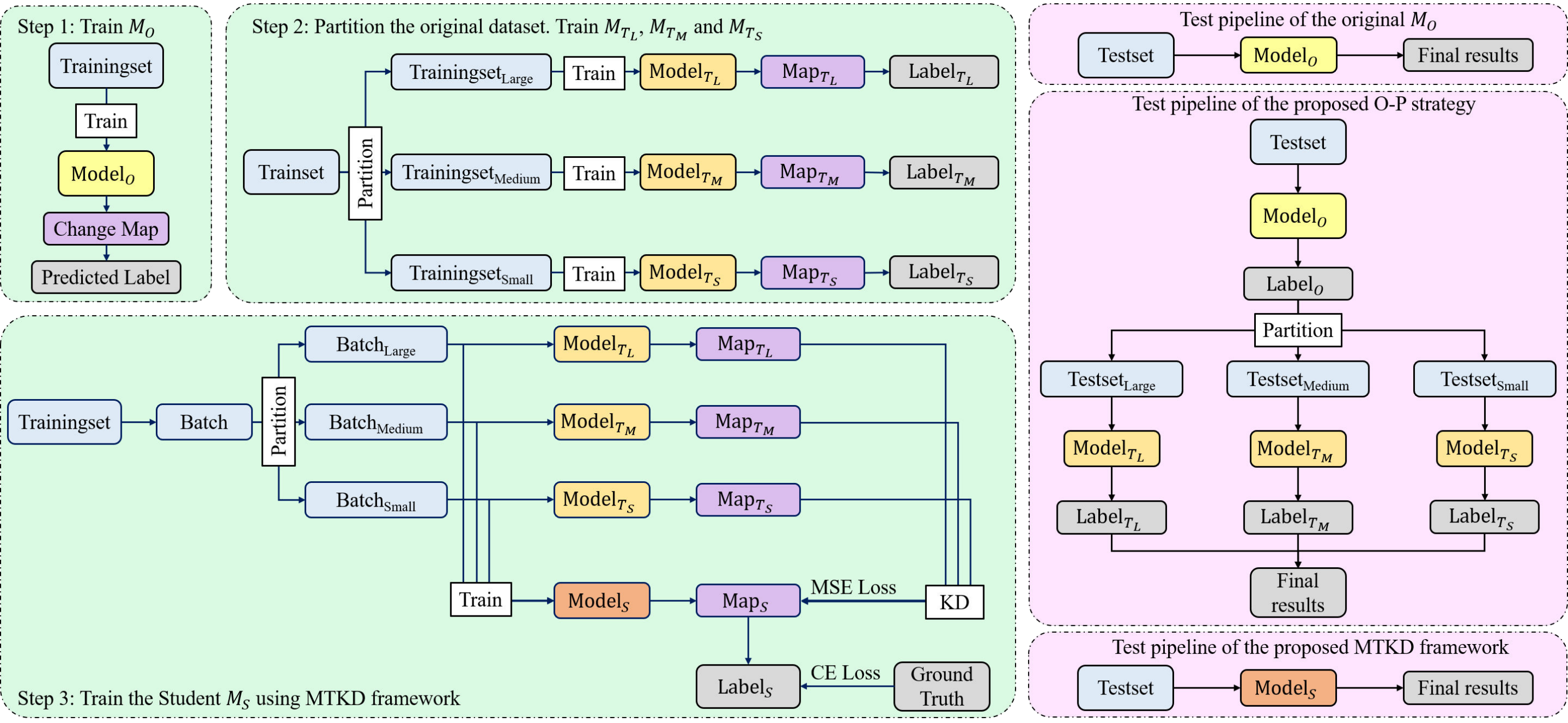}
    \caption{Overview of the training (green boxes) and testing (pink boxes) pipelines of the proposed Origin-Partition (O-P) strategy and Multi-Teacher Knowledge Distillation (MTKD) framework.}
    \label{fig:pipeline}
\end{figure*}

\section{Methodology} \label{Methodology}
In this section, we provide a comprehensive overview of the proposed methods. 
In Section \ref{O-P Strategy}, we first introduce the Origin-Partition (O-P) strategy designed for the challenging all-inclusive CD dataset. 
Building upon the O-P strategy, we further present our Multi-Teacher Knowledge Distillation (MTKD) framework in Section \ref{MTKD Framework}. 
Finally, in Section \ref{Loss Function}, we describe the overall loss function used for training the teacher and student models.

\subsection{O-P Strategy} \label{O-P Strategy}
The traditional training and testing paradigm for CD models is depicted in the upper-left (training, green box) and upper-right (testing, pink box) sections of Figure \ref{fig:pipeline}.
In binary CD tasks, given a model $\mathcal{M}$ and an image pair $(X_1, X_2)$, the model outputs a change map (CM), from which the predicted label $\hat{Y}$ is derived via thresholding or element-wise comparison.

To quantify the diversity of changes within a dataset, we introduce the concept of Change Area Ratio (CAR), defined as the ratio of changed pixels relative to the total image area. 
A detailed discussion of the CAR distribution on the JL1-CD and SYSU-CD datasets is provided in Appendix~\ref{section: car distribution}.
For datasets like them, where CAR values span a wide range (0\% to 100\%), traditional single-stage training tends to yield models that struggle to perform well across all CAR scenarios.
To address this issue, we propose a general Origin-Partition (O-P) strategy.
Rather than training a single model on the entire training set, we partition the dataset into multiple subsets according to different CAR thresholds, which enables each model to focus on a specific range of change scenarios.
As a concrete instantiation (see Fig.~\ref{fig:pipeline}, green boxes at Step 1 and Step 2), for a given CD algorithm, we train the corresponding models in the following sequence:

1) The original model $\mathcal{M}_O$ is trained on the complete training set using the algorithm's default configuration.

2) To reduce the training complexity, we set appropriate CAR thresholds $th_1$ and $th_2$ to partition the original training set into three categories: small, medium, and large.

3) The models are then trained from scratch using the partitioned training sets, yielding models $\mathcal{M}_{T_S}$, $\mathcal{M}_{T_M}$, and $\mathcal{M}_{T_L}$. The training process can be formalized as:
\begin{equation}
\hat{Y} = 
\begin{cases}
f_{\mathcal{M}_{T_S}}(X_1, X_2), & \text{if } \text{CAR}_\text{GT} \leq th_1; \\
f_{\mathcal{M}_{T_M}}(X_1, X_2), & \text{if } th_1 < \text{CAR}_\text{GT} \leq th_2; \\
f_{\mathcal{M}_{T_L}}(X_1, X_2), & \text{if } \text{CAR}_\text{GT} > th_2,
\end{cases}
\end{equation}
where $\text{CAR}_\text{GT}$ denotes the CAR calculated based on the ground truth label for the image pair $(X_1, X_2)$.

As shown in the middle pink box in Fig. \ref{fig:pipeline}, during testing, since the true CAR of the test images is unknown, we first use $\mathcal{M}_O$ to estimate the CAR roughly. 
Based on this estimated CAR, we then assign the images to one of the three categories: small, medium, or large, and send them to the corresponding model $\mathcal{M}_{T_S}$, $\mathcal{M}_{T_M}$, or $\mathcal{M}_{T_L}$ to obtain the final detection result:
\begin{equation}
\hat{Y} = 
\begin{cases}
f_{\mathcal{M}_{T_S}}(X_1, X_2), & \text{if } \text{CAR}_{\mathcal{M}_O} \leq th_1; \\
f_{\mathcal{M}_{T_M}}(X_1, X_2), & \text{if } th_1 < \text{CAR}_{\mathcal{M}_O} \leq th_2; \\
f_{\mathcal{M}_{T_L}}(X_1, X_2), & \text{if } \text{CAR}_{\mathcal{M}_O} > th_2,
\end{cases}
\end{equation}
where $\text{CAR}_{\mathcal{M}_O}$ denotes the CAR calculated based on the predicted label from the original model $\mathcal{M}_O$.

\subsection{MTKD Framework} \label{MTKD Framework}

Partitioning the training set in the O-P strategy effectively reduces the training complexity for each model and can enhance CD performance to some extent.
Nevertheless, this approach suffers from two major limitations:
\begin{enumerate}[1)]
    \item During inference, we are required to load four different models, and even disregarding data throughput, the time required is at least twice that of the original method, significantly increasing both memory and computational complexity.
    \item Since we first use $\mathcal{M}_O$ to obtain a rough estimate of the CAR, any inaccuracies in this estimation may lead to incorrect model selection in the subsequent steps, thereby reducing the accuracy of the final predictions.
\end{enumerate}
These motivate us to explore whether it is possible to integrate the strengths of the original model and those trained on sub-datasets into a single model.

To address these limitations, we propose the Multi-Teacher Knowledge Distillation (MTKD) framework, as illustrated in the green box at the bottom of Fig.~\ref{fig:pipeline}. 
In the O-P strategy, we have already trained the models $\mathcal{M}_O$, $\mathcal{M}_{T_S}$, $\mathcal{M}_{T_M}$, and $\mathcal{M}_{T_L}$. 
Building on this, we further train a student model $\mathcal{M}_S$. 
First, we initialize the student model $\mathcal{M}_S$ using the parameters from $\mathcal{M}_O$, and then use $\mathcal{M}_{T_S}$, $\mathcal{M}_{T_M}$, and $\mathcal{M}_{T_L}$ as teacher models to perform KD. 
For each input image pair, we select the appropriate teacher model according to its CAR to supervise the student model. 
In this framework, the student model $\mathcal{M}_S$ is simultaneously supervised by the ground truth labels and the CM information from the teacher models across different CAR partitions.

During the testing phase, only the student model $\mathcal{M}_S$ is used for inference, thereby improving the model’s CD performance across different CAR ranges without introducing any additional computational cost.

\begin{table*}[!t]
\centering
\caption{Benchmark Methods and the Corresponding Implementation Details\label{table:benchmark}}
\renewcommand{\arraystretch}{1.25}
\begin{tabular}{lcccccccc}
\toprule
Method & Backbone & Param (M) & Flops (G) & Initial LR & $\lambda$ & Scheduler & Batch Size & GPU  \\ \hline
FC-EF\cite{fcsn} & \cellcolor[HTML]{ECF4FF}CNN 
& 1.353 & 12.976 & 1e-4 & 1e-4 & LinearLR & 8 & 3090 \\
FC-Siam-Conc\cite{fcsn} & \cellcolor[HTML]{ECF4FF}CNN 
& 1.548 & 19.956 & 1e-4 & 1e-4 & LinearLR & 8 & 3090 \\
FC-Siam-Diff\cite{fcsn} & \cellcolor[HTML]{ECF4FF}CNN 
& 1.352 & 17.540 & 1e-4 & 1e-4 & LinearLR & 8 & 3090 \\ \hline
STANet\cite{stanet} & \cellcolor[HTML]{ECF4FF}ResNet-18  
& 12.764 & 70.311 & 1e-3 & 5e-3 & LinearLR & 8 & 3090 \\ \hline
IFN\cite{ifn} & \cellcolor[HTML]{ECF4FF}VGG-16 
& 35.995 & 323.584 & 1e-3 & 1e-4 & LinearLR & 8 & 3090 \\ \hline
SNUNet\cite{snunet} & \cellcolor[HTML]{ECF4FF}CNN  
& 3.012 & 46.921 & 1e-3 & 1e-4 & LinearLR & 8 & 3090 \\ \hline
BIT\cite{bit} & \cellcolor[HTML]{ECF4FF}ResNet-18           
& 2.990 & 34.996 & 1e-3 & 1e-4 & LinearLR & 8 & 3090 \\ \hline
ChangeStar-FarSeg\cite{changestar} & \cellcolor[HTML]{ECF4FF}ResNet-18  
& 16.965 & 76.845 & 1e-3 & 1e-4 & LinearLR & 16 & 3090 \\
ChangeStar-UPerNet\cite{changestar} & \cellcolor[HTML]{ECF4FF}ResNet-18 
& 13.952 & 55.634 & 1e-3 & 1e-4 & LinearLR & 8 & 3090 \\ \hline 
& \cellcolor[HTML]{96FFFB}MiT-b0              
& 3.847  & 11.380 & 6e-5 & 1e-4 & LinearLR & 8 & 3090 \\
\multirow{-2}{*}{ChangeFormer\cite{changeformer}} 
& \cellcolor[HTML]{96FFFB}MiT-b1              
& 13.941 & 26.422 & 6e-5 & 5e-4 & LinearLR & 8 & 3090 \\ \hline
TinyCD\cite{tinycd} & \cellcolor[HTML]{ECF4FF}CNN 
& 0.285 & 5.791 & 3.57e-3 & 1e-5 & LinearLR & 8 & 3090 \\ \hline
HANet\cite{hanet} & \cellcolor[HTML]{ECF4FF}CNN                 
& 3.028 & 97.548 & 1e-3 & 1e-3 & LinearLR & 8 & A800 \\ \hline
& \cellcolor[HTML]{96FFFB}MiT-b0              
& 3.457 & 8.523  & 1e-4 & 1e-4 & LinearLR & 8 & 3090 \\ 
& \cellcolor[HTML]{96FFFB}MiT-b1              
& 13.355 & 23.306 & 1e-4 & 1e-3 & LinearLR & 8 & 3090 \\ 
& \cellcolor[HTML]{ECF4FF}ResNet-18           
& 11.391 & 23.820 & 5e-3 & 1e-3 & LinearLR & 8 & 3090 \\
\multirow{-4}{*}{Changer\cite{changer}} & \cellcolor[HTML]{ECF4FF}ResNeSt-50          
& 26.693 & 67.241 & 5e-3 & 1e-5 & LinearLR & 8 & 3090 \\ \hline
LightCDNet\cite{lightcdnet} & \cellcolor[HTML]{ECF4FF}CNN                 
& 0.342 & 6.995 & 3e-3 & 5e-3 & LinearLR & 8 & 3090 \\ \hline
CGNet\cite{cgnet} & \cellcolor[HTML]{ECF4FF}VGG-16   
& 38.989 & 425.984 & 5e-4 & 1e-3 & LinearLR & 8 & A800 \\ \hline
BAN\cite{ban} & \cellcolor[HTML]{CBCEFB}ViT-L 
& 261.120 & 346.112 & 1e-4 & 1e-3 & LinearLR & 8 & A800 \\ \hline
TTP\cite{ttp} & \cellcolor[HTML]{CBCEFB}SAM\cite{sam} 
& 361.472 & 929.792 & 4e-4 & 5e-3 & CosineAnnealingLR & 8 & A800 \\ \hline
\end{tabular}
\end{table*}

\subsection{Loss Function} \label{Loss Function}
For training the original model $\mathcal{M}_O$ as well as the teacher models $\mathcal{M}_{T_S}$, $\mathcal{M}_{T_M}$, and $\mathcal{M}_{T_L}$, we employ the standard binary cross-entropy loss:
\begin{equation}
\mathcal{L}_{\text{CE}} = \frac{1}{N} \sum_{i=1}^{N} ( -Y(i) \log(\hat{Y}(i)) - (1 - Y(i)) \log(1 - \hat{Y}(i)) ),
\end{equation}
where $Y(i)$ denotes the ground truth label of pixel $i$.

During the training of the student model $\mathcal{M}_S$, we dynamically select the appropriate teacher model according to the corresponding CAR range.
The knowledge distillation loss is defined as the mean squared error (MSE) between the outputs of the student and the selected teacher at the CM layer:
\begin{equation}
\mathcal{L}_{\text{KD}} = \frac{1}{N} \sum_{i=1}^{N} \left( \text{CM}_S(i) - \text{CM}_{\mathcal{T}}(i) \right)^2, \quad \mathcal{T} \in \{ T_S, T_M, T_L \},
\end{equation}
where $\text{CM}_S(i)$ and $\text{CM}_{\mathcal{T}}(i)$ denote the outputs at the CM layer for the student and the selected teacher, respectively.
The overall loss for training $\mathcal{M}_S$ is a weighted sum of the above objectives:
\begin{equation}
\mathcal{L} = \mathcal{L}_{\text{CE}} + \lambda \mathcal{L}_{\text{KD}},
\end{equation}
where $\lambda$ is used to balance the cross-entropy loss and the distillation loss.

\begin{table*}[!t]
\centering
\caption{Experimental Results on JL1-CD Test Set \label{table:metrics}}
\renewcommand{\arraystretch}{1.25}
\begin{tabular}{ccccccccccccc}
\toprule
Method & Strategy & mIoU & mRec & mPrec & mFscore &  & Method & Strategy & mIoU  & mRec & mPrec & mFscore \\
\cellcolor[HTML]{EFEFEF} & \cellcolor[HTML]{EFEFEF}-  
& \cellcolor[HTML]{EFEFEF}57.08 
& \cellcolor[HTML]{EFEFEF}61.90 
& \cellcolor[HTML]{EFEFEF}86.40 
& \cellcolor[HTML]{EFEFEF}61.28 
&  & \cellcolor[HTML]{EFEFEF} & \cellcolor[HTML]{EFEFEF}- 
& \cellcolor[HTML]{EFEFEF}63.79 
& \cellcolor[HTML]{EFEFEF}69.54 
& \cellcolor[HTML]{EFEFEF}84.77 
& \cellcolor[HTML]{EFEFEF}69.19 \\
\cellcolor[HTML]{EFEFEF} & \cellcolor[HTML]{EFEFEF}O-P  
& \cellcolor[HTML]{EFEFEF}{50.05 }
& \cellcolor[HTML]{EFEFEF}{53.64 }
& \cellcolor[HTML]{EFEFEF}{\textbf{95.28}}
& \cellcolor[HTML]{EFEFEF}{52.00 }
&  & \cellcolor[HTML]{EFEFEF}  & \cellcolor[HTML]{EFEFEF}O-P  
& \cellcolor[HTML]{EFEFEF}{60.93} 
& \cellcolor[HTML]{EFEFEF}{64.72} 
& \cellcolor[HTML]{EFEFEF}{\textbf{91.03} }
& \cellcolor[HTML]{EFEFEF}{65.46} \\
\multirow{-3}{*}{\cellcolor[HTML]{EFEFEF}FC-EF}  
& \cellcolor[HTML]{EFEFEF}MTKD 
& \cellcolor[HTML]{EFEFEF}{\textbf{61.87} }
& \cellcolor[HTML]{EFEFEF}{\textbf{68.45} }
& \cellcolor[HTML]{EFEFEF}{81.90} 
& \cellcolor[HTML]{EFEFEF}{\textbf{67.33} }
&  & \multirow{-3}{*}{\cellcolor[HTML]{EFEFEF}FC-Siam-Conc}                                                    & \cellcolor[HTML]{EFEFEF}MTKD 
& \cellcolor[HTML]{EFEFEF}{\textbf{64.30} }
& \cellcolor[HTML]{EFEFEF}{\textbf{70.58} }
& \cellcolor[HTML]{EFEFEF}{83.76 }
& \cellcolor[HTML]{EFEFEF}{\textbf{69.88}} \\
& - & 61.30 & 66.03 & 86.45 & 66.34 &  &  
& -  & 66.76 & 81.71 & 74.73 & 74.02 \\
& O-P & {60.93}  & {64.72} & {\textbf{91.03}} & {65.46} &  &  
& O-P & {64.56} & {78.47} & {72.13} & {71.25} \\
\multirow{-3}{*}{FC-Siam-Diff} & MTKD 
& {\textbf{63.75}} & {\textbf{69.57}} & {83.95} & {\textbf{69.49}} &  & 
\multirow{-3}{*}{STANet} & MTKD 
& {\textbf{67.92}}  & {\textbf{82.07}}  & {\textbf{76.24}} & {\textbf{75.10}} \\
\cellcolor[HTML]{EFEFEF} & \cellcolor[HTML]{EFEFEF}-    
& \cellcolor[HTML]{EFEFEF}71.25          
& \cellcolor[HTML]{EFEFEF}78.91          
& \cellcolor[HTML]{EFEFEF}84.53          
& \cellcolor[HTML]{EFEFEF}77.33          
&  & \cellcolor[HTML]{EFEFEF} & \cellcolor[HTML]{EFEFEF}-   
& \cellcolor[HTML]{EFEFEF}68.97          
& \cellcolor[HTML]{EFEFEF}74.87          
& \cellcolor[HTML]{EFEFEF}\textbf{85.06} 
& \cellcolor[HTML]{EFEFEF}75.25  \\
\cellcolor[HTML]{EFEFEF} & \cellcolor[HTML]{EFEFEF}O-P  
& \cellcolor[HTML]{EFEFEF}{71.06}          
& \cellcolor[HTML]{EFEFEF}{78.37}          
& \cellcolor[HTML]{EFEFEF}{84.28}          
& \cellcolor[HTML]{EFEFEF}{77.21}          
&  & \cellcolor[HTML]{EFEFEF} & \cellcolor[HTML]{EFEFEF}O-P  
& \cellcolor[HTML]{EFEFEF}{\textbf{71.39} }
& \cellcolor[HTML]{EFEFEF}{\textbf{78.60} }
& \cellcolor[HTML]{EFEFEF}{83.36}          
& \cellcolor[HTML]{EFEFEF}{\textbf{77.98}} \\
\multirow{-3}{*}{\cellcolor[HTML]{EFEFEF}IFN}                                                            & \cellcolor[HTML]{EFEFEF}MTKD & 
\cellcolor[HTML]{EFEFEF}{\textbf{72.72}} 
& \cellcolor[HTML]{EFEFEF}{\textbf{80.28} }
& \cellcolor[HTML]{EFEFEF}{\textbf{84.66} }
& \cellcolor[HTML]{EFEFEF}{\textbf{78.80} }
&  & \multirow{-3}{*}{\cellcolor[HTML]{EFEFEF}SNUNet} & \cellcolor[HTML]{EFEFEF}MTKD 
& \cellcolor[HTML]{EFEFEF}{71.12 }
& \cellcolor[HTML]{EFEFEF}{78.27}
& \cellcolor[HTML]{EFEFEF}{84.96   }
& \cellcolor[HTML]{EFEFEF}{77.56}  \\
& - & 67.22 & 74.47 & 83.71 & 73.37 &  & 
& - & 69.47 & 75.58 & 84.46 & 75.57 \\
& O-P  & {\textbf{69.41}} & {\textbf{76.29}} & {84.02} & {\textbf{75.77}} &  &  
& O-P  & {68.87} & {74.74} & {84.90} & {74.86} \\
\multirow{-3}{*}{BIT} & MTKD 
& {68.86} & {75.49} & {\textbf{84.71}} & {74.88} &  
& \multirow{-3}{*}{\begin{tabular}[c]{@{}c@{}}ChangeStar\\ (FarSeg)\end{tabular}}  & MTKD 
& {\textbf{69.87}} & {\textbf{75.80}} & {\textbf{85.78}} & {\textbf{76.03}} \\
\cellcolor[HTML]{EFEFEF} & \cellcolor[HTML]{EFEFEF}-  
& \cellcolor[HTML]{EFEFEF}64.85          
& \cellcolor[HTML]{EFEFEF}69.18          
& \cellcolor[HTML]{EFEFEF}\textbf{88.26} 
& \cellcolor[HTML]{EFEFEF}70.19          
&  & \cellcolor[HTML]{EFEFEF} & \cellcolor[HTML]{EFEFEF}-    
& \cellcolor[HTML]{EFEFEF}73.51          
& \cellcolor[HTML]{EFEFEF}\textbf{80.46} 
& \cellcolor[HTML]{EFEFEF}{86.33} 
& \cellcolor[HTML]{EFEFEF}{79.70} \\
\cellcolor[HTML]{EFEFEF}  & \cellcolor[HTML]{EFEFEF}O-P  
& \cellcolor[HTML]{EFEFEF}{64.68} 
& \cellcolor[HTML]{EFEFEF}{69.05}
& \cellcolor[HTML]{EFEFEF}{87.23} 
& \cellcolor[HTML]{EFEFEF}{70.08} 
&  & \cellcolor[HTML]{EFEFEF} & \cellcolor[HTML]{EFEFEF}O-P  
& \cellcolor[HTML]{EFEFEF}{72.58}          
& \cellcolor[HTML]{EFEFEF}{79.16}          
& \cellcolor[HTML]{EFEFEF}{86.33}
& \cellcolor[HTML]{EFEFEF}{78.79}   \\
\multirow{-3}{*}{\cellcolor[HTML]{EFEFEF}\begin{tabular}[c]{@{}c@{}}ChangeStar\\ (UPerNet)\end{tabular}} & \cellcolor[HTML]{EFEFEF}MTKD 
& \cellcolor[HTML]{EFEFEF}{\textbf{65.10} }
& \cellcolor[HTML]{EFEFEF}{\textbf{70.26}} 
& \cellcolor[HTML]{EFEFEF}{87.69}          
& \cellcolor[HTML]{EFEFEF}{\textbf{70.58}} 
&  & \multirow{-3}{*}{\cellcolor[HTML]{EFEFEF}\begin{tabular}[c]{@{}c@{}}ChangeFormer\\ (MiT-b0)\end{tabular}} & \cellcolor[HTML]{EFEFEF}MTKD 
& \cellcolor[HTML]{EFEFEF}{\textbf{73.78}} 
& \cellcolor[HTML]{EFEFEF}{80.07 }         
& \cellcolor[HTML]{EFEFEF}\textbf{86.90 }         
& \cellcolor[HTML]{EFEFEF}\textbf{79.98} \\ 
& - & 73.05 & 79.70 & 86.95  & 79.22  & & & - & 71.04 & 78.77 & 83.05 & 77.74 \\
& O-P & {73.45} & {79.19} & {\textbf{87.46}} & {79.41} &  &  
& O-P & {72.22} & {79.93} & {\textbf{83.49}} & {78.76} \\ 
\multirow{-3}{*}{\begin{tabular}[c]{@{}c@{}}ChangeFormer\\ (MiT-b1)\end{tabular}} & MTKD 
& {\textbf{73.92}} & {\textbf{80.43}} & {86.89} & {\textbf{80.18}} 
&  & \multirow{-3}{*}{TinyCD} & MTKD 
& {\textbf{72.55}}  & {\textbf{80.98}} & {83.17} & {\textbf{79.26}} \\
\cellcolor[HTML]{EFEFEF} & \cellcolor[HTML]{EFEFEF}-    
& \cellcolor[HTML]{EFEFEF}63.64          
& \cellcolor[HTML]{EFEFEF}69.77          
& \cellcolor[HTML]{EFEFEF}83.43          
& \cellcolor[HTML]{EFEFEF}69.39          
&  & \cellcolor[HTML]{EFEFEF} & \cellcolor[HTML]{EFEFEF}-    
& \cellcolor[HTML]{EFEFEF}74.85          
& \cellcolor[HTML]{EFEFEF}\textbf{81.84} 
& \cellcolor[HTML]{EFEFEF}86.09          
& \cellcolor[HTML]{EFEFEF}80.98          \\
\cellcolor[HTML]{EFEFEF}  & \cellcolor[HTML]{EFEFEF}O-P  
& \cellcolor[HTML]{EFEFEF}{\textbf{69.05} }
& \cellcolor[HTML]{EFEFEF}{\textbf{76.53} }
& \cellcolor[HTML]{EFEFEF}{83.05  }        
& \cellcolor[HTML]{EFEFEF}{\textbf{75.66} }
&  & \cellcolor[HTML]{EFEFEF} & \cellcolor[HTML]{EFEFEF}O-P  
& \cellcolor[HTML]{EFEFEF}{75.29 }         
& \cellcolor[HTML]{EFEFEF}{81.40 }         
& \cellcolor[HTML]{EFEFEF}{87.06}          
& \cellcolor[HTML]{EFEFEF}{\textbf{81.32}} \\
\multirow{-3}{*}{\cellcolor[HTML]{EFEFEF}HANet} & \cellcolor[HTML]{EFEFEF}MTKD 
& \cellcolor[HTML]{EFEFEF}{67.67}          
& \cellcolor[HTML]{EFEFEF}{74.39}          
& \cellcolor[HTML]{EFEFEF}{\textbf{84.38} }
& \cellcolor[HTML]{EFEFEF}{73.92 }         
&  & \multirow{-3}{*}{\cellcolor[HTML]{EFEFEF}\begin{tabular}[c]{@{}c@{}}Changer\\ (MiT-b0)\end{tabular}}      & \cellcolor[HTML]{EFEFEF}MTKD 
& \cellcolor[HTML]{EFEFEF}{\textbf{75.35} }
& \cellcolor[HTML]{EFEFEF}{81.76 }         
& \cellcolor[HTML]{EFEFEF}{\textbf{87.18}} 
& \cellcolor[HTML]{EFEFEF}{81.28}  \\
& - & 75.94 & 81.99 & \textbf{87.74} & 81.93 &  &  & - & 68.37 & 75.15 & 83.43 & 74.54  \\ 
& O-P & {75.42} & {81.67} & {87.13}  & {81.43} & & 
& O-P & {\textbf{70.76}} & {\textbf{77.42}} & {\textbf{83.86}} & {\textbf{77.01}} \\
\multirow{-3}{*}{\begin{tabular}[c]{@{}c@{}}Changer\\ (MiT-b1)\end{tabular}} & MTKD 
& {\textbf{76.15}} & {\textbf{82.85}} & {86.98} & {\textbf{82.13}} &  
& \multirow{-3}{*}{\begin{tabular}[c]{@{}c@{}}Changer\\ (r18)\end{tabular}} & MTKD 
& {69.45} & {77.26} & {81.50} & {75.86}  \\
\cellcolor[HTML]{EFEFEF} & \cellcolor[HTML]{EFEFEF}-    
& \cellcolor[HTML]{EFEFEF}62.31          
& \cellcolor[HTML]{EFEFEF}69.23          
& \cellcolor[HTML]{EFEFEF}80.91          
& \cellcolor[HTML]{EFEFEF}67.83          
&  & \cellcolor[HTML]{EFEFEF} & \cellcolor[HTML]{EFEFEF}-    
& \cellcolor[HTML]{EFEFEF}64.58         
& \cellcolor[HTML]{EFEFEF}71.16          
& \cellcolor[HTML]{EFEFEF}83.12          
& \cellcolor[HTML]{EFEFEF}70.13  \\
\cellcolor[HTML]{EFEFEF}  & \cellcolor[HTML]{EFEFEF}O-P  
& \cellcolor[HTML]{EFEFEF}{\textbf{71.80} }
& \cellcolor[HTML]{EFEFEF}{\textbf{79.76} }
& \cellcolor[HTML]{EFEFEF}{\textbf{83.15} }
& \cellcolor[HTML]{EFEFEF}{\textbf{78.23} }
&  & \cellcolor[HTML]{EFEFEF}  & \cellcolor[HTML]{EFEFEF}O-P  
& \cellcolor[HTML]{EFEFEF}{\textbf{70.19} }
& \cellcolor[HTML]{EFEFEF}{\textbf{77.43} }
& \cellcolor[HTML]{EFEFEF}{\textbf{83.99}} 
& \cellcolor[HTML]{EFEFEF}{\textbf{76.16}} \\
\multirow{-3}{*}{\cellcolor[HTML]{EFEFEF}\begin{tabular}[c]{@{}c@{}}Changer\\ (s50)\end{tabular}}  & \cellcolor[HTML]{EFEFEF}MTKD 
& \cellcolor[HTML]{EFEFEF}{62.96} 
& \cellcolor[HTML]{EFEFEF}{69.65} 
& \cellcolor[HTML]{EFEFEF}{81.76} 
& \cellcolor[HTML]{EFEFEF}{68.52} 
&  & \multirow{-3}{*}{\cellcolor[HTML]{EFEFEF}LightCDNet} & \cellcolor[HTML]{EFEFEF}MTKD 
& \cellcolor[HTML]{EFEFEF}{65.99}          
& \cellcolor[HTML]{EFEFEF}{72.44}          
& \cellcolor[HTML]{EFEFEF}{83.86}          
& \cellcolor[HTML]{EFEFEF}{71.48}  \\ 
& -   & 73.37  & 80.31  & 85.33 & 79.65 &  &  & -  & 73.54  & 79.54 & 87.89 & 79.47  \\ 
& O-P & {72.95} & {79.71} & {85.50} & {79.12} &  &  
& O-P & {73.61} & {79.17} & {\textbf{88.10}} & {79.45} \\
\multirow{-3}{*}{CGNet} & MTKD 
& {\textbf{73.82}} & {\textbf{80.32}} &{ \textbf{86.33}} & {\textbf{79.91}} 
&  & \multirow{-3}{*}{BAN}  & MTKD  
& {\textbf{73.95}}  & {\textbf{80.26}}  & {87.12} & {\textbf{79.92}} \\
\cellcolor[HTML]{EFEFEF} & \cellcolor[HTML]{EFEFEF}-    
& \cellcolor[HTML]{EFEFEF}75.05          
& \cellcolor[HTML]{EFEFEF}80.24          
& \cellcolor[HTML]{EFEFEF}\textbf{89.82} 
& \cellcolor[HTML]{EFEFEF}80.76  & & & & & & & \\
\cellcolor[HTML]{EFEFEF} & \cellcolor[HTML]{EFEFEF}O-P  
& \cellcolor[HTML]{EFEFEF}{76.69}          
& \cellcolor[HTML]{EFEFEF}{\textbf{83.48}} 
& \cellcolor[HTML]{EFEFEF}{87.27}         
& \cellcolor[HTML]{EFEFEF}{82.52} 
& & & & & & & \\
\cellcolor[HTML]{EFEFEF}                                                                                 \multirow{-3}{*}{\cellcolor[HTML]{EFEFEF}\begin{tabular}[c]{@{}c@{}}TTP\end{tabular}} & \cellcolor[HTML]{EFEFEF}MTKD 
& \cellcolor[HTML]{EFEFEF}{\textbf{76.85}} 
& \cellcolor[HTML]{EFEFEF}{82.99 }         
& \cellcolor[HTML]{EFEFEF}{88.05}          
& \cellcolor[HTML]{EFEFEF}{\textbf{82.56}} 
& & & & & & & \\ \hline
\end{tabular}
\end{table*}

\begin{figure*}[!t]
    \centering
        \includegraphics[
        width=1.0\textwidth,
        trim=0.5cm 0 0 0, 
        clip
    ]{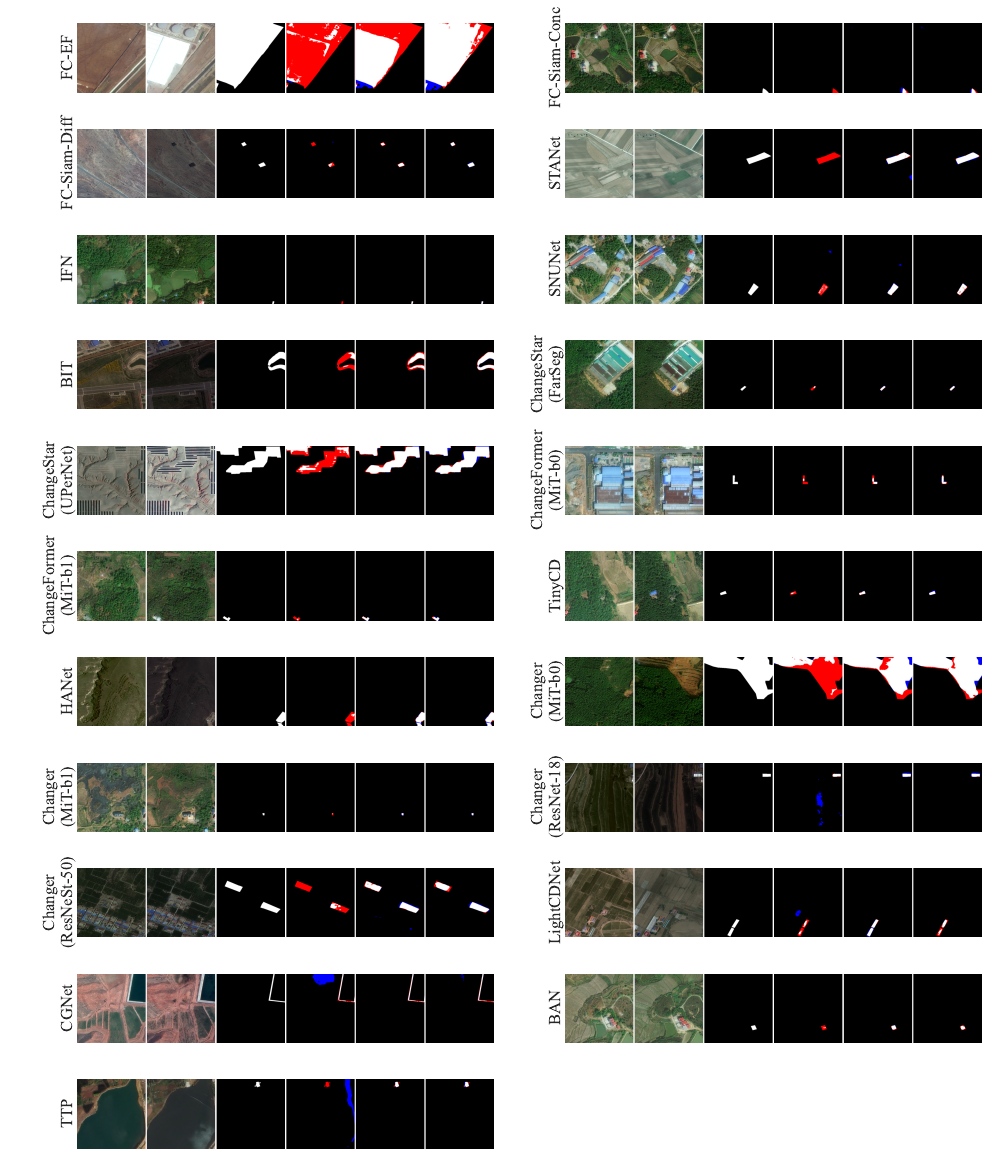}
    \caption{Visual comparison on the JL1-CD dataset. For each method, the images from left to right are: pre-event image, post-event image, ground truth, output of the original model, output of the O-P strategy, and output of the MTKD framework. Red indicates missed detections, while blue represents false alarms.}
    \label{fig:visual-v2}
\end{figure*}

\begin{figure*}[!t]
    \centering
    \subfloat[FC-EF]{\includegraphics[width=0.22\linewidth]{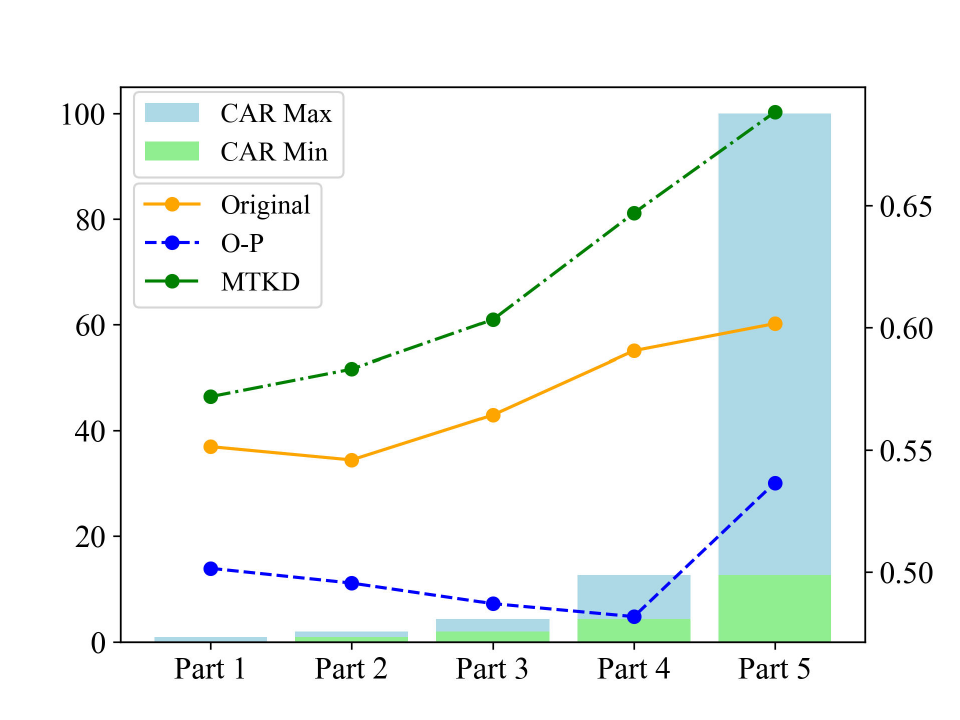}}
    \label{fig:CAR-Partition-FC-EF}
    \subfloat[FC-Siam-Conc]{\includegraphics[width=0.22\linewidth]{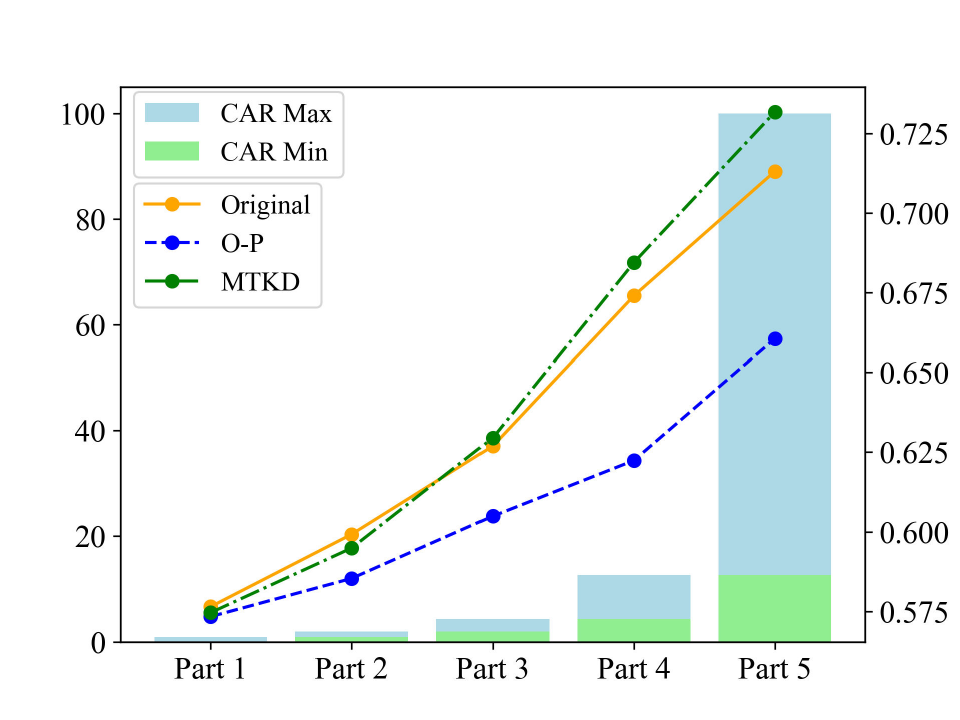}}
    \label{fig:CAR-Partition-FC-Siam-Conc}
    \subfloat[FC-Siam-Diff]{\includegraphics[width=0.22\linewidth]{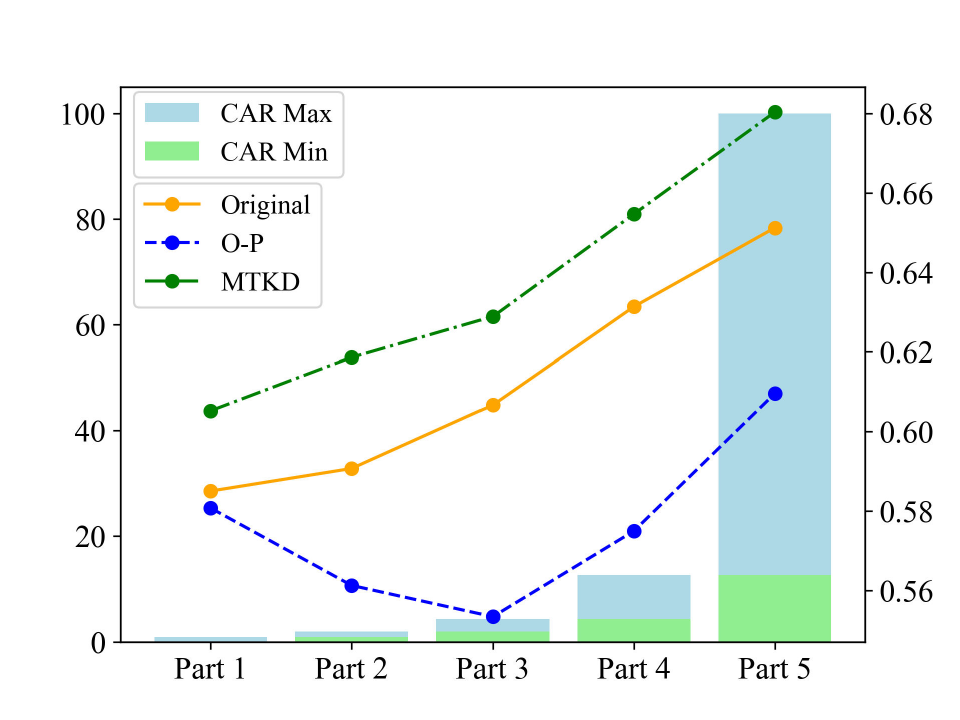}}
    \label{fig:CAR-Partition-FC-Siam-Diff}
    \subfloat[STANet]{\includegraphics[width=0.22\linewidth]{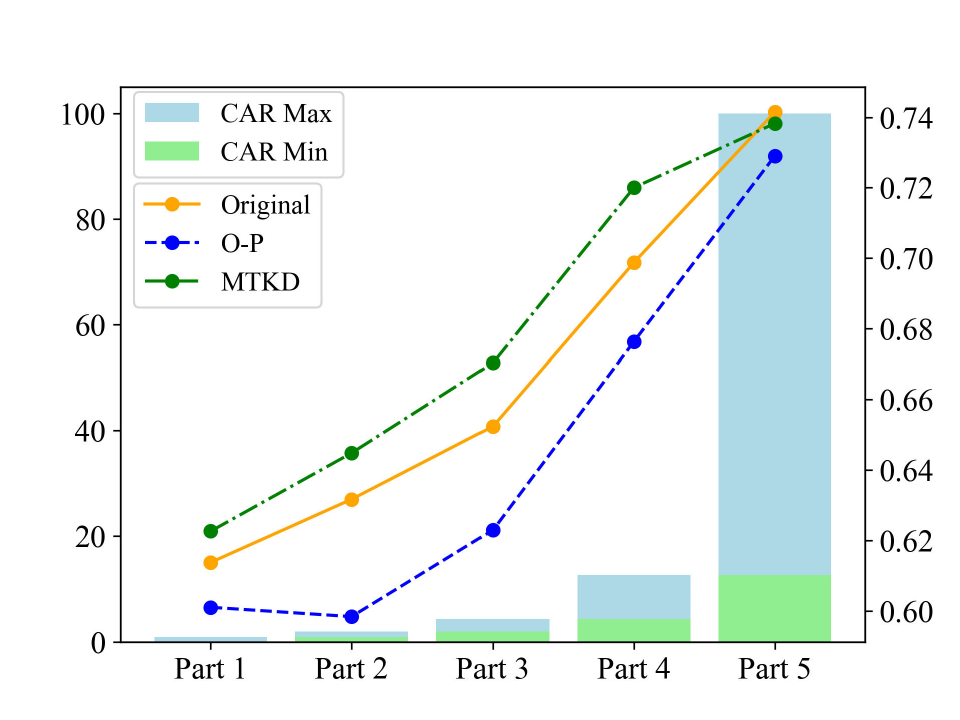}}
    \label{fig:CAR-Partition-STANet}
    \\
    \subfloat[IFN]{\includegraphics[width=0.22\linewidth]{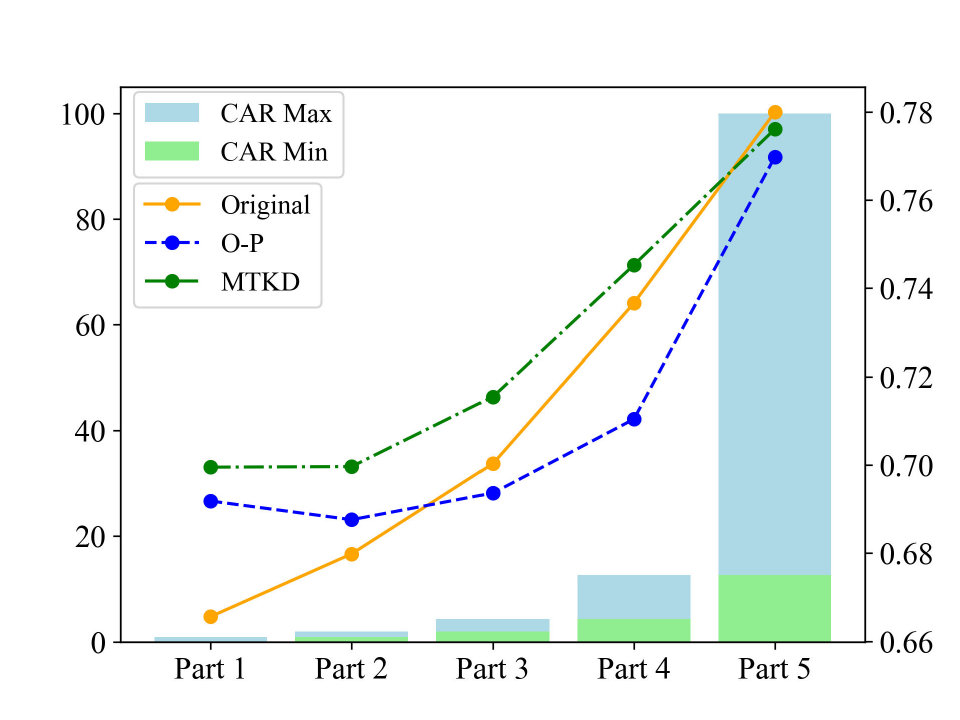}}
    \label{fig:CAR-Partition-IFN}
    \subfloat[SNUNet]{\includegraphics[width=0.22\linewidth]{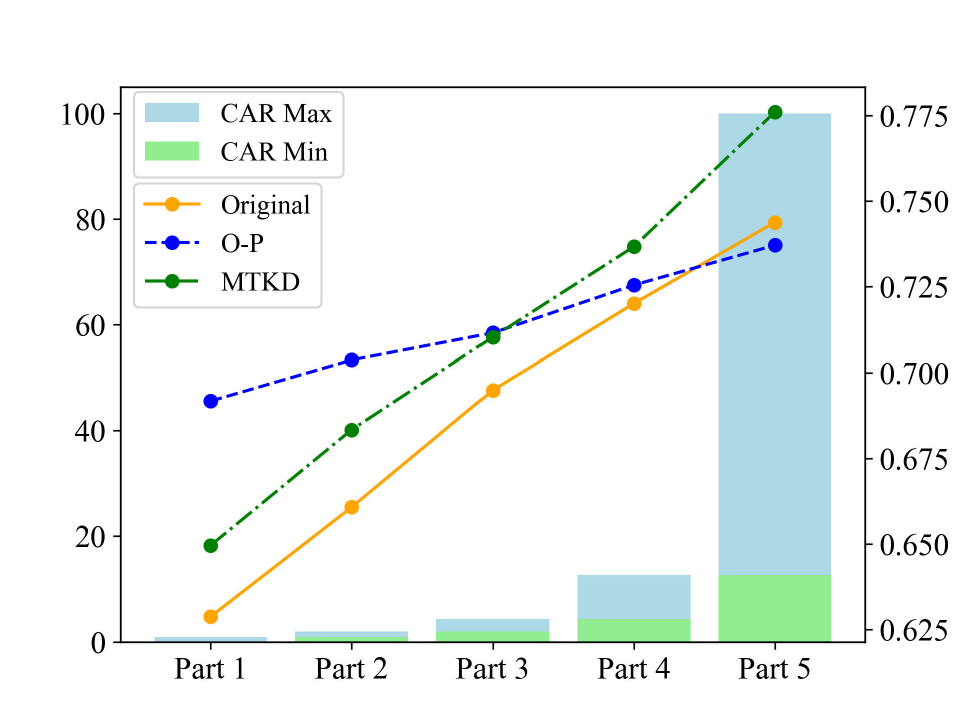}}
    \label{fig:CAR-Partition-SNUNet}
    \subfloat[BIT]{\includegraphics[width=0.22\linewidth]{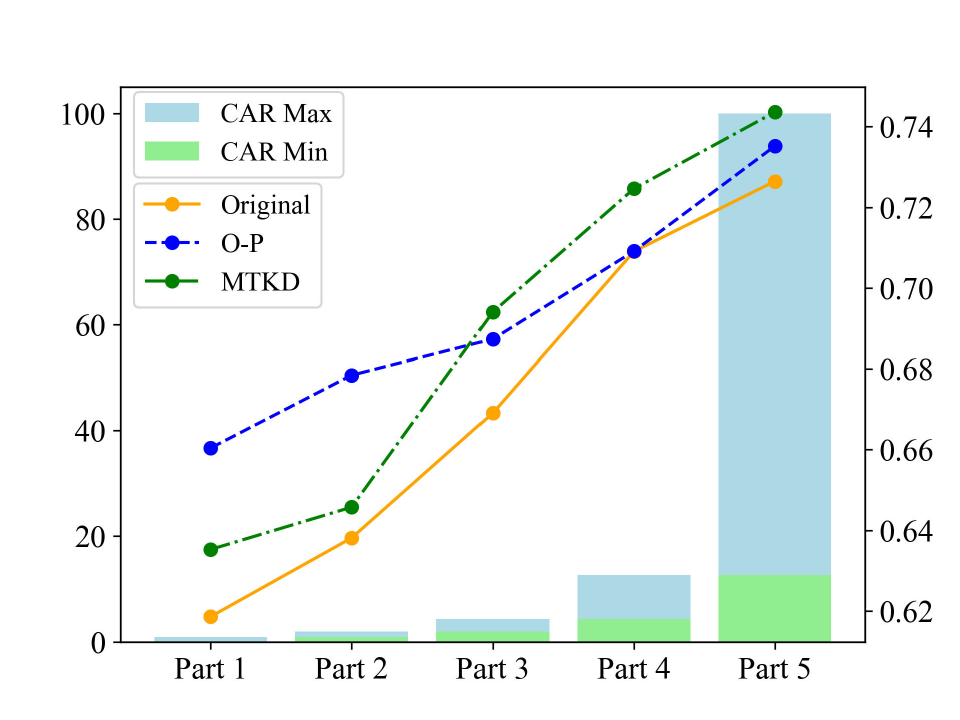}}
    \label{fig:CAR-Partition-BIT}
    \subfloat[ChangeStar (FarSeg)]{\includegraphics[width=0.22\linewidth]{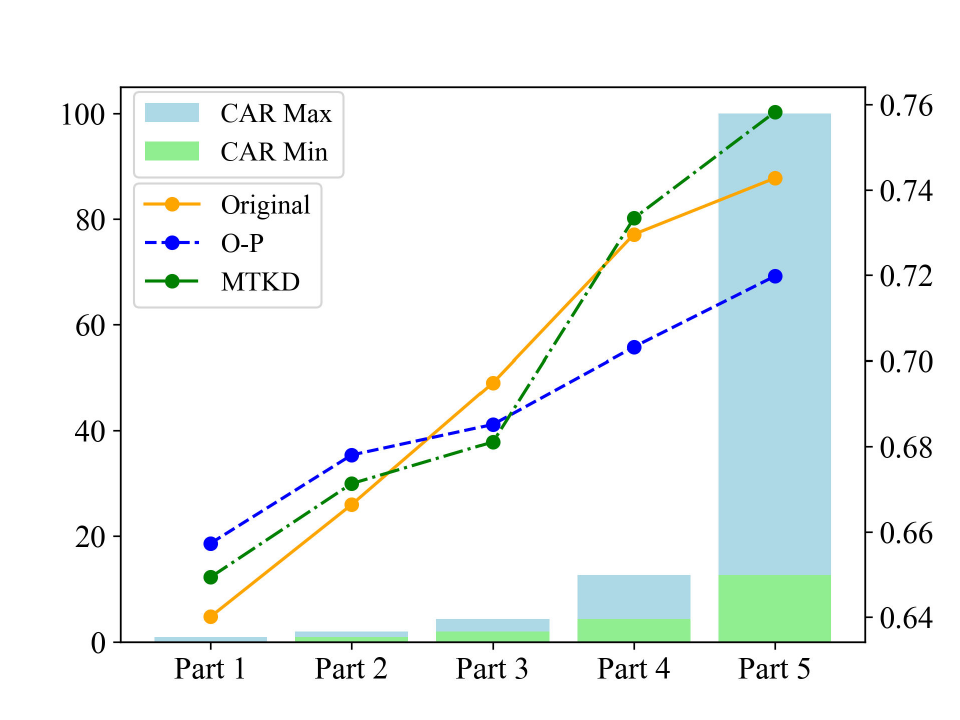}}
    \label{fig:CAR-Partition-ChangeStar-FarSeg}
    \\
    \subfloat[ChangeStar (UPerNet)]{\includegraphics[width=0.22\linewidth]{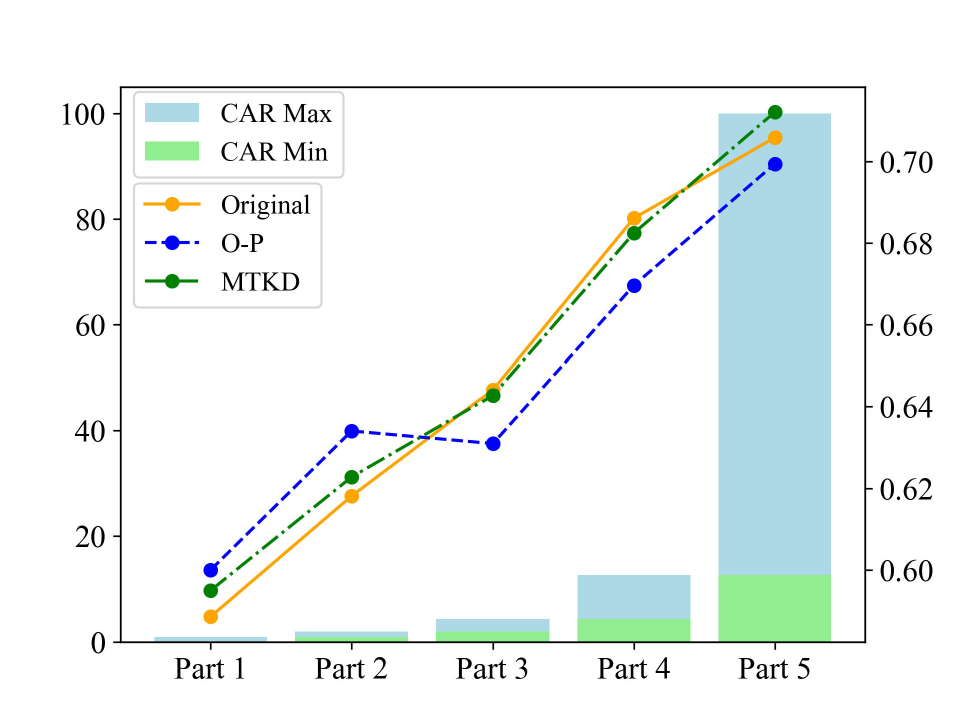}}
    \label{fig:CAR-Partition-ChangeStar-UPerNet}
    \subfloat[ChangeFormer (MiT-b0)]{\includegraphics[width=0.22\linewidth]{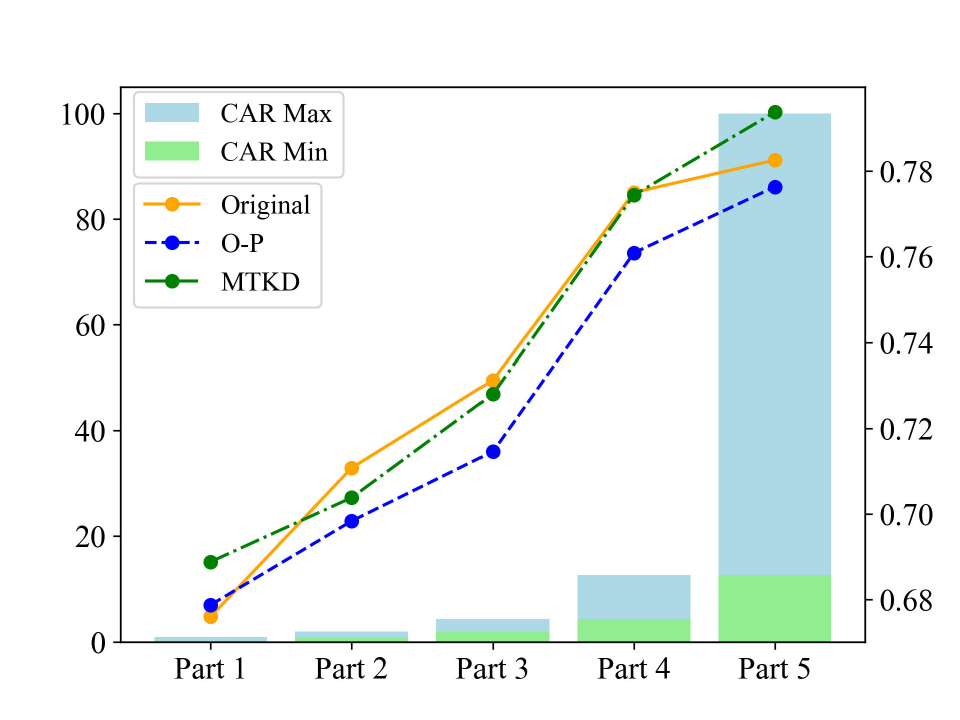}}
    \label{fig:CAR-Partition-ChangeFormer-MiT-b0}
    \subfloat[ChangeFormer (MiT-b1)]{\includegraphics[width=0.22\linewidth]{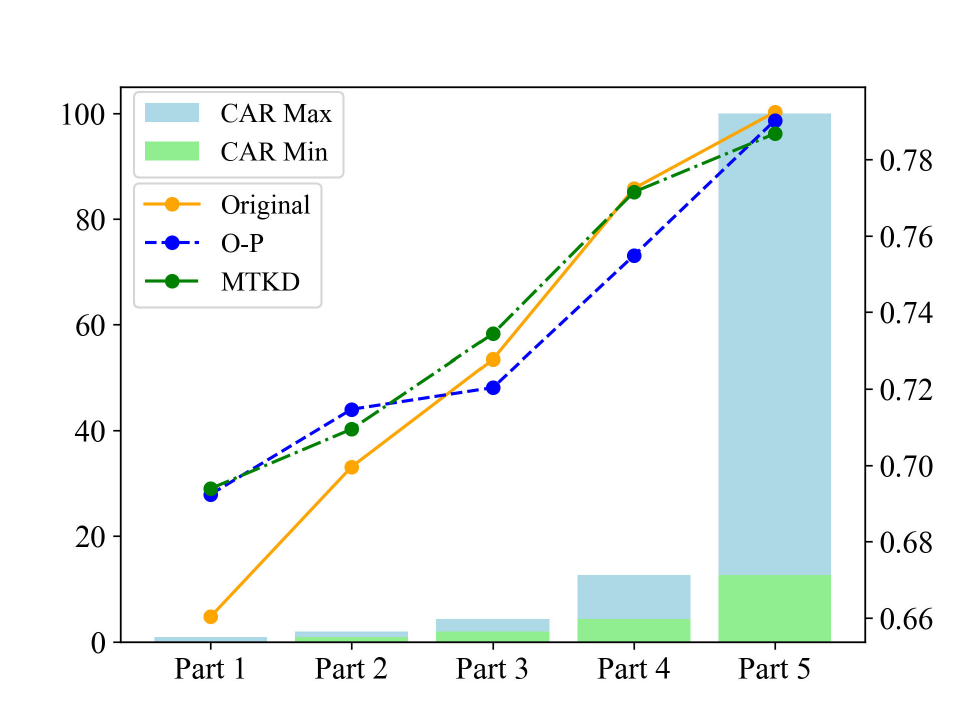}}
    \label{fig:CAR-Partition-ChangeFormer-MiT-b1}
    \subfloat[TinyCD]{\includegraphics[width=0.22\linewidth]{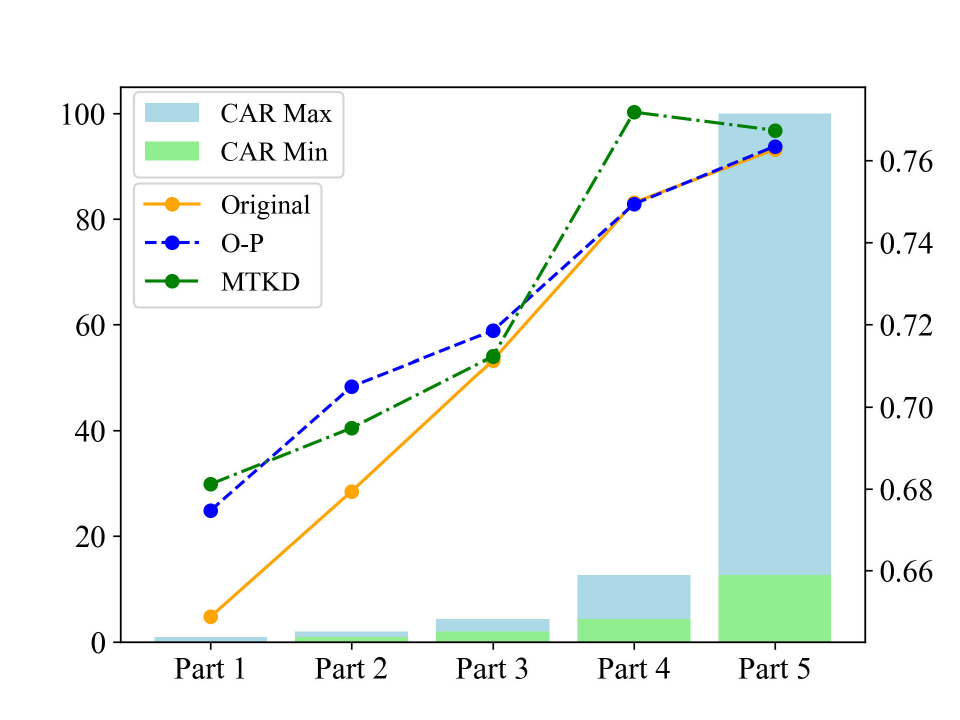}}
    \label{fig:CAR-Partition-TinyCD}
    \\
    \subfloat[HANet]{\includegraphics[width=0.22\linewidth]{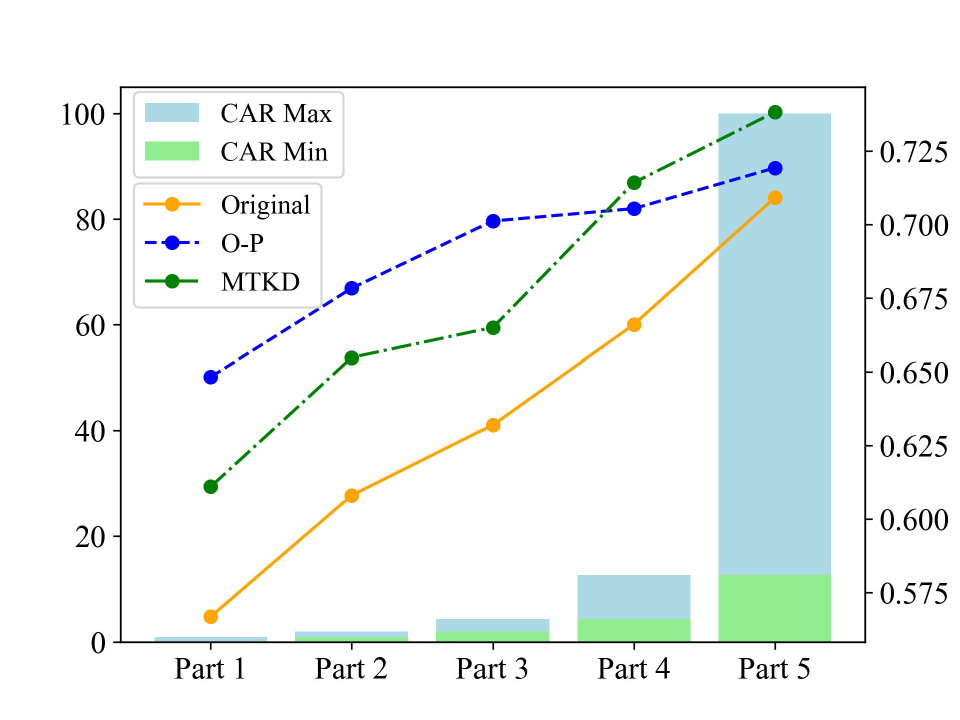}}
    \label{fig:CAR-Partition-HANet}
    \subfloat[Changer (MiT-b0)]{\includegraphics[width=0.22\linewidth]{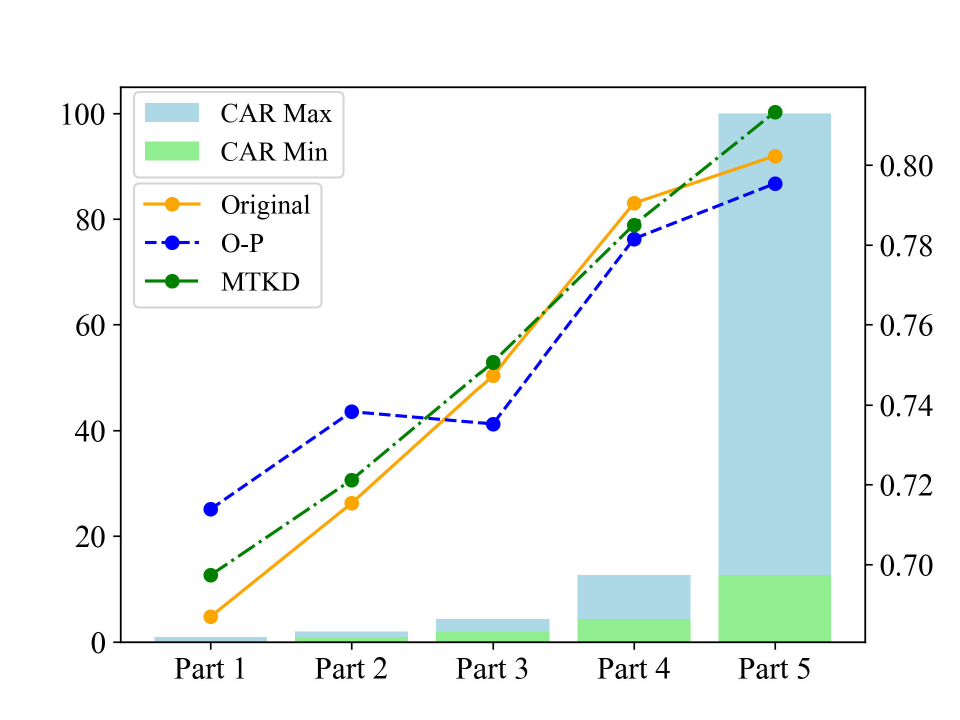}}
    \label{fig:CAR-Partition-Changer-MiT-b0}
    \subfloat[Changer (MiT-b1)]{\includegraphics[width=0.22\linewidth]{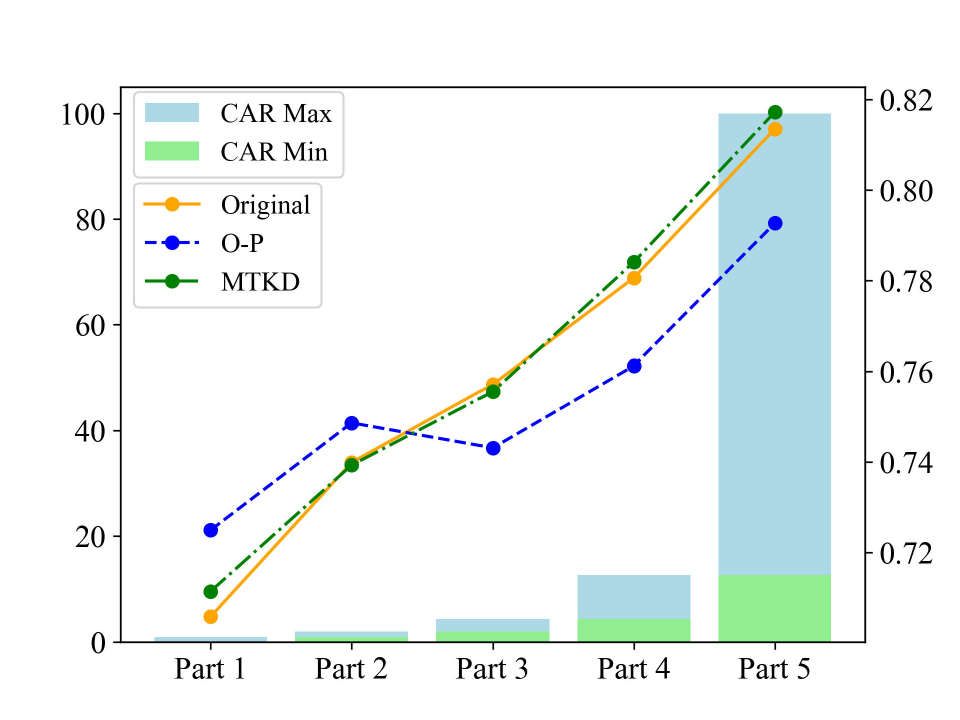}}
    \label{fig:CAR-Partition-Changer-MiT-b1}
    \subfloat[Changer (ResNet-18)]{\includegraphics[width=0.22\linewidth]{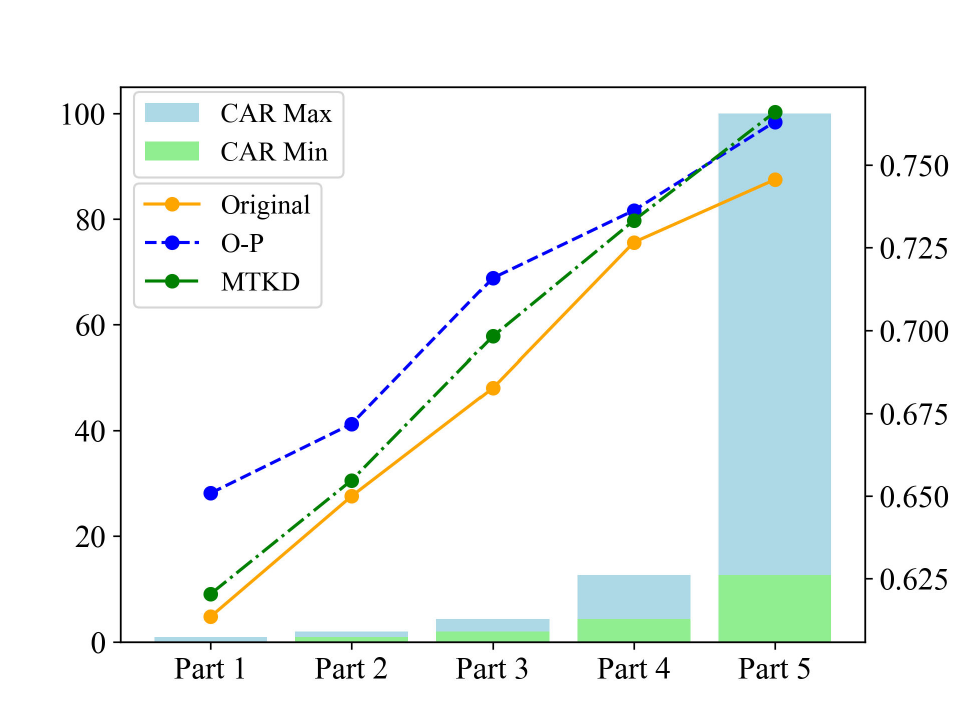}}
    \label{fig:CAR-Partition-Changer-r18}
    \\
    \subfloat[Changer (ResNeSt-50)]{\includegraphics[width=0.22\linewidth]{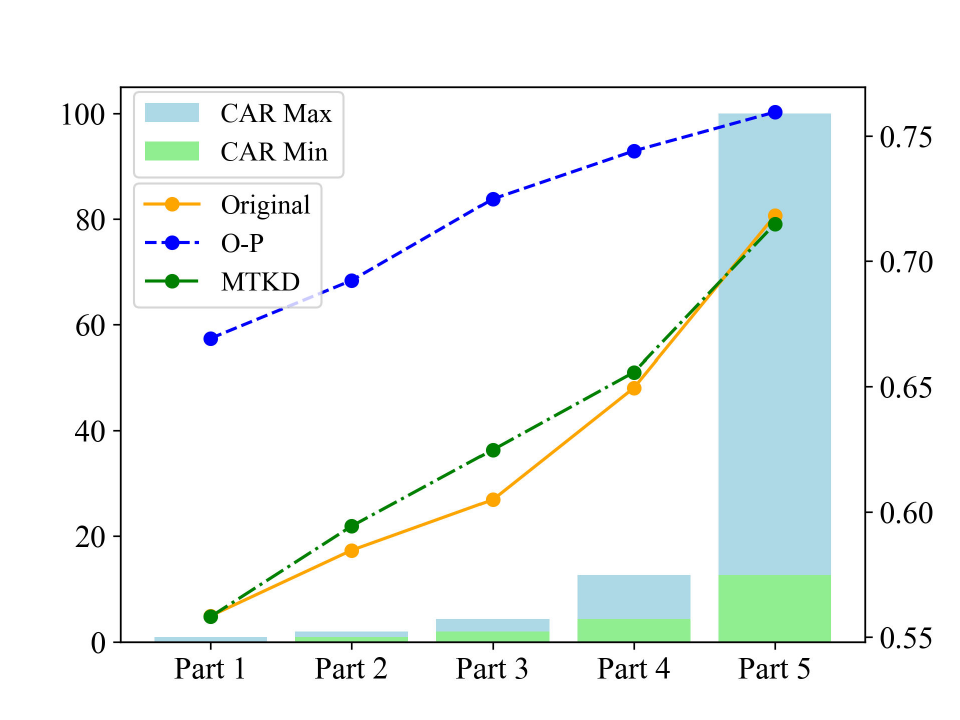}}
    \label{fig:CAR-Partition-Changer-s50}
    \subfloat[LightCDNet]{\includegraphics[width=0.22\linewidth]{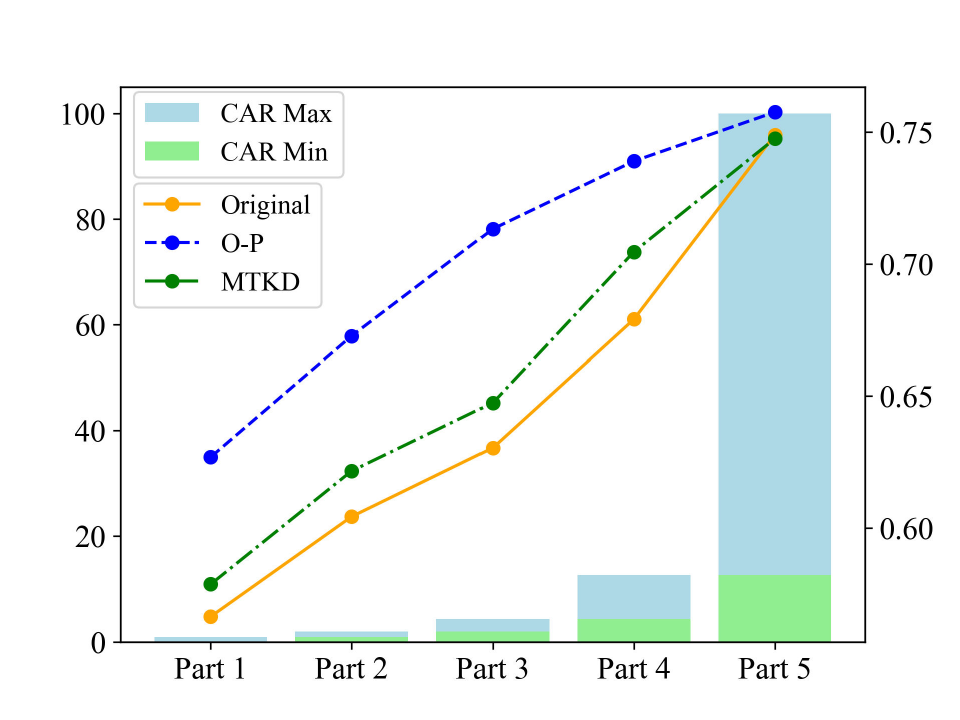}}
    \label{fig:CAR-Partition-LightCDNet}
    \subfloat[CGNet]{\includegraphics[width=0.22\linewidth]{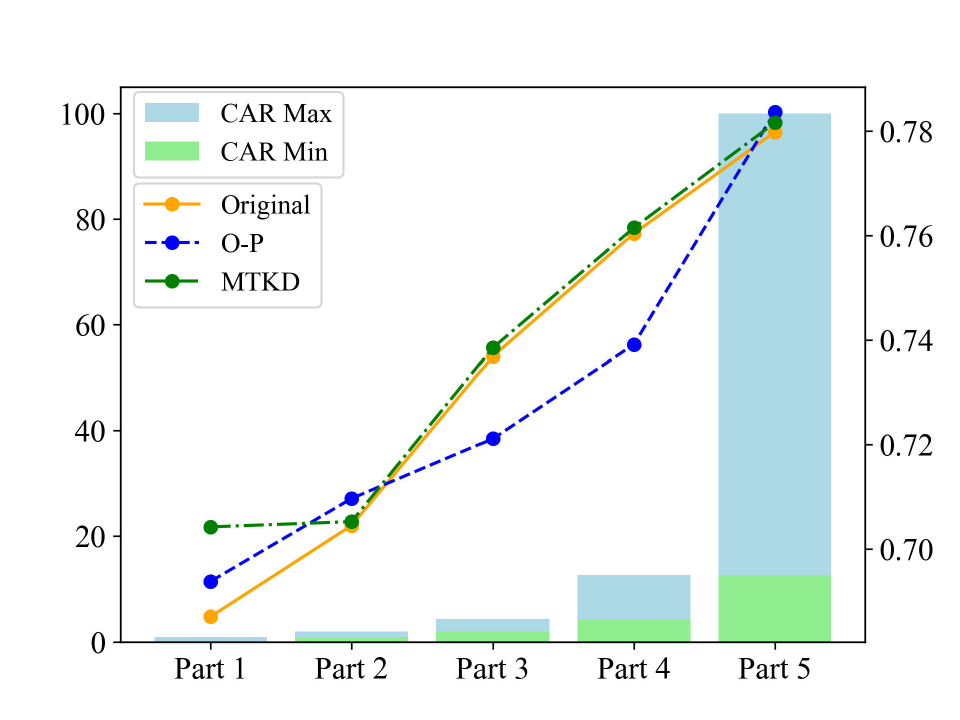}}
    \label{fig:CAR-Partition-CGNet}
    \subfloat[BAN]{\includegraphics[width=0.22\linewidth]{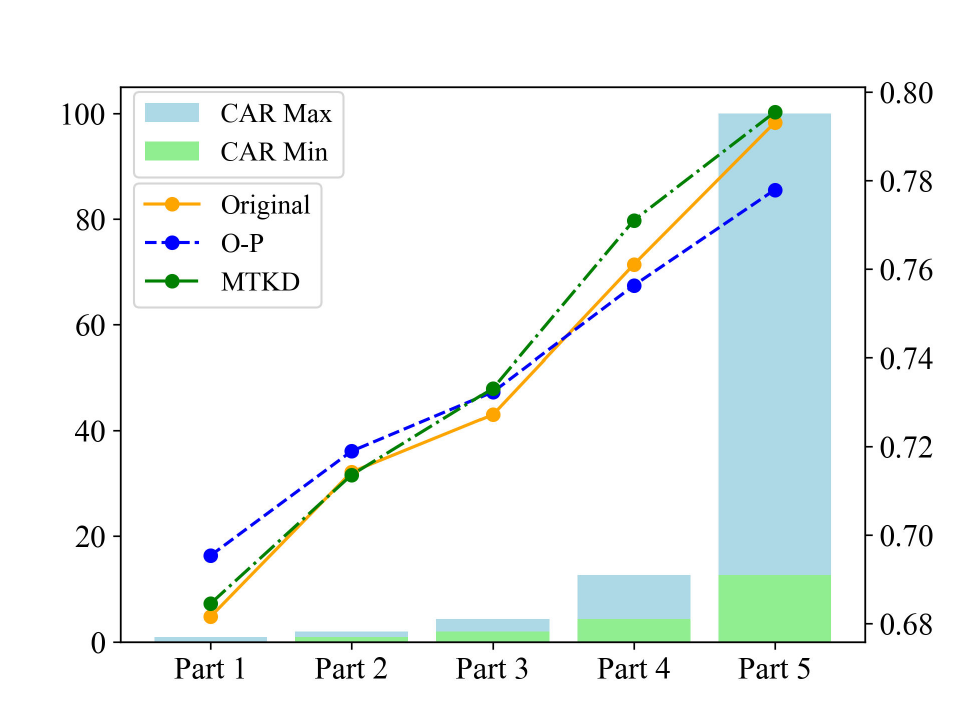}}
    \label{fig:CAR-Partition-BAN}
    \\
    \subfloat[TTP]{\includegraphics[width=0.22\linewidth]{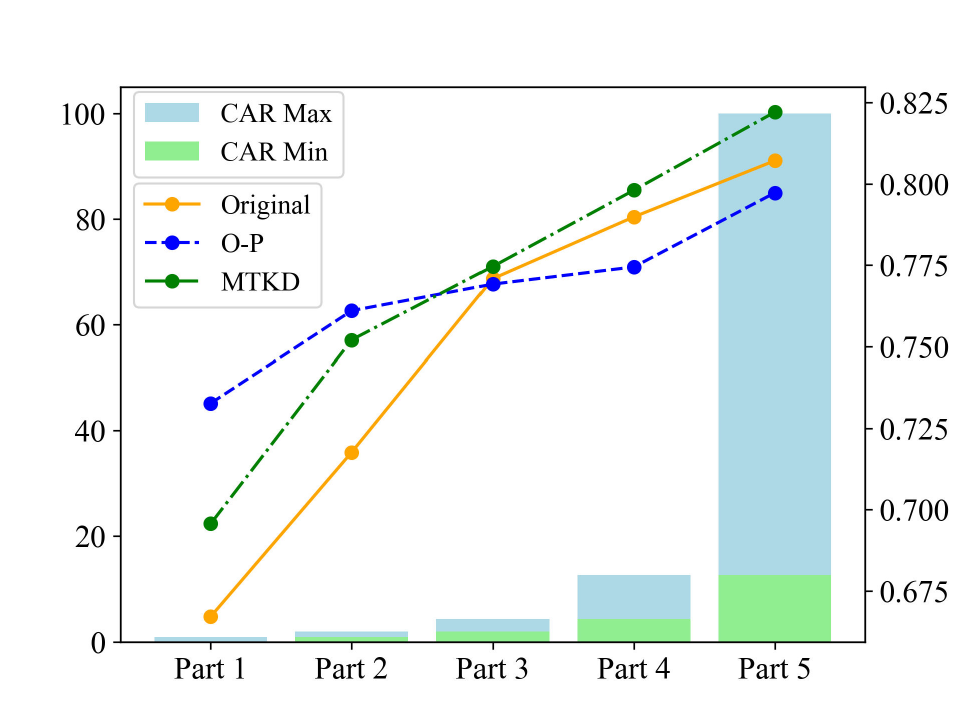}}
    \label{fig:CAR-Partition-TTP}
    \caption{mIoU of the benchmark CD methods across different CAR ranges. In each plot, the left y-axis represents CAR size, and the right y-axis represents mIoU.}
    \label{fig:car_per_partition}
\end{figure*}

\section{Experiment} \label{Experiment}      

\subsection{Dataset Description} \label{Dataset Description}
We first conduct experiments on our JL1-CD dataset. 
To validate the robustness of the proposed MTKD framework, we further perform experiments on the SYSU-CD dataset \cite{sysucd}. The detailed information for these two datasets is as follows:

1) JL1-CD Dataset: As described in Section \ref{JL1-CD Dataset}, the JL1-CD dataset consists of 5,000 pairs of high-resolution images, with a resolution of 0.5-0.75 meters and image size of 512 × 512. In the competition, the first 4,000 image pairs are used for training, and the remaining 1,000 image pairs are used for testing. 

2) SYSU-CD Dataset: As illustrated in Fig. \ref{fig:CAR-SYSU-CD}, the SYSU-CD dataset is also a challenging dataset with a very large CAR range, so we choose this dataset as the second benchmark to validate the robustness of our proposed methods. 
The SYSU-CD dataset contains 12,000 pairs for training, 4,000 pairs for validation, and 4,000 pairs for testing, with each image having a size of 256 × 256.
\subsection{Benchmark Methods} \label{Benchmark Methods}
To comprehensively verify the validity of our JL1-CD dataset and the effectiveness of the proposed methods, we conduct extensive experiments on existing benchmark algorithms. 
The selected models, along with their corresponding backbones, parameter sizes, and computational complexities, are summarized in Table \ref{table:benchmark}. 
Based on the backbone architecture, the models are categorized into three groups: 
Alice Blue for CNN-based models, Light Cyan for Transformer-based models, and Lavender Blue for FM-based models, encompassing almost all mainstream architectures. 
FLOPs is calculated with an input image of size 512 × 512. As shown in the table, the selected models span a wide range of sizes, from lightweight models with fewer than 1M parameters, such as TinyCD and LightCDNet, to the latest SOTA model TTP with over 360M parameters.
This wide range of model sizes allows us to verify the universality of the O-P and MTKD methods across models with different backbones and scales.

\subsection{Evaluation Metrics} \label{Evaluation Metrics}
The common evaluation metrics for CD models include Intersection over Union (IoU), Precision (Prec), Recall (Rec), and F1-score (Fscore). 
IoU measures the overlap between the detected change region and the ground truth. 
Accuracy reflects the overall correctness of the model. 
Precision indicates the false positive rate of the model, Recall reflects the false negative rate, and F1-score balances both of them. A higher F1-score indicates better detection performance.
Given the large CAR range in the JL1-CD dataset, both change and non-change regions are equally important. Therefore, we choose the averaged versions of the aforementioned metrics, which are calculated as follows:
\begin{equation}
\begin{aligned}
\text{mIoU} &= \frac{1}{2} \left( \text{IoU}_0 + \text{IoU}_1 \right) \\
&= \frac{1}{2} \left( \frac{\text{TN}}{\text{TN} + \text{FP} + \text{FN}} + \frac{\text{TP}}{\text{TP} + \text{FP} + \text{FN}} \right),
\end{aligned}
\end{equation}
\begin{equation}
\begin{aligned}
\text{mPrec} &= \frac{1}{2} \left( \text{Prec}_0 + \text{Prec}_1 \right) \\
&= \frac{1}{2} \left( \frac{\text{TN}}{\text{TN} + \text{FN}} + \frac{\text{TP}}{\text{TP} + \text{FP}} \right) ,
\end{aligned}
\end{equation}
\begin{equation}
\begin{aligned}
\text{mRec} &= \frac{1}{2} \left( \text{Rec}_0 + \text{Rec}_1 \right) \\
&= \frac{1}{2} \left( \frac{\text{TN}}{\text{TN} + \text{FP}} + \frac{\text{TP}}{\text{TP} + \text{FN}} \right) ,
\end{aligned}
\end{equation}
\begin{equation}
\begin{aligned}
\text{mFscore} &= \frac{1}{2} \left( \text{Fscore}_0 + \text{Fscore}_1 \right) \\
&= \frac{1}{2}\sum_{i=0}^1\frac{2 \times \text{Prec}_i \times \text{Rec}_i}{\text{Prec}_i + \text{Rec}_i},
\end{aligned}
\end{equation}
where TP, FP, TN, and FN represent true positives, false positives, true negatives, and false negatives, respectively. 
To align with the ``Jilin-1'' Cup competition requirements, we first calculate these metrics for each individual image and then compute the average of the results across all images to obtain the final outcome.

\subsection{Implementation Details} \label{Implementation Details}
All models are trained and evaluated with the OpenCD Toolbox built on PyTorch \cite{opencd}. 
To ensure fair comparisons and to clearly assess the contributions of the Origin-Partition (O-P) strategy and the Multi-Teacher Knowledge Distillation (MTKD) framework to the performance improvement of the CD models, we adopt consistent parameter settings and hardware conditions for all models ($\mathcal{M}_O$, $\mathcal{M}_{T_L}$, $\mathcal{M}_{T_M}$, $\mathcal{M}_{T_S}$, and $\mathcal{M}_S$) across the various algorithms, as described below:

\begin{enumerate}[1)]
    \item The patch size of the input images is 512 × 512, which matches the original image dimensions. 
    \item Data augmentation methods, including RandomRotate, RandomFlip, and PhotoMetricDistortion, are applied.
    \item The AdamW optimizer is used with $\beta_1 = 0.9$ and $\beta_2 = 0.99$, and the default batch size is set to 8 image pairs (with a batch size of 16 for ChangeStar-FarSeg).
    \item Models $\mathcal{M}_O$, $\mathcal{M}_{T_L}$, $\mathcal{M}_{T_M}$, and $\mathcal{M}_{T_S}$ are trained for 200k iterations on the original and the corresponding partitioned datasets (300 epochs for TTP), while the student model $\mathcal{M}_S$ is trained for an additional 100k iterations (100 epochs for TTP) on the original dataset.
    \item Training begins with a warm-up phase of 1k iterations (5 epochs for TTP), during which the learning rate (LR) is linearly increased from 1e-6 to the initial LR value, as specified in Table \ref{table:benchmark}. 
Afterward, the LR is linearly decayed to 0 as training progresses (TTP employs a CosineAnnealing decay schedule).
    \item The HANet, CGNet, BAN, and TTP models are trained on the NVIDIA A800 server, while other models are trained on the NVIDIA RTX 3090 server. All models are tested on the A800 server.
\end{enumerate}

When dividing the training set into three partitions, we empirically set the thresholds as $th_1 = 0.05$ and $th_2 = 0.2$. For the case using only two teacher models, the threshold is set to $0.10$.
The model is saved every 1k iterations (5 epochs for TTP), and the checkpoint with the highest mIoU value on the validation set is selected for testing.

For the training of $\mathcal{M}_{T_S}$, 
due to the varying distributions of the change maps and the different magnitudes of $\mathcal{L}_{\text{CE}}$ across models, a grid search is performed to determine the optimal distillation loss weight $\lambda$ for each model. 
More configuration details are summarized in Table \ref{table:benchmark}.

\subsection{Experimental Results} \label{Experimental Results}
\subsubsection{Quantitative Comparison}  
Table \ref{table:metrics} summarizes the numerical results of mIoU, mRecall (mRec), mPrecision (mPrec), and mF1-score (mFscore) for all methods on the JL1-CD test set, trained under the original, O-P, and MTKD strategies. 
As shown in the table, the MTKD framework can improve the performance of all the algorithms to a certain extent. 
After MTKD optimization, the Changer-MiT-b0 and Changer-MiT-b1 models can outperform the original TTP in terms of mIoU. Additionally, the TTP model, after MTKD optimization, shows improvements in mIoU and mFscore by 1.30\% and 1.80\%, respectively, setting a new SOTA.
However, several algorithms, such as STANet, IFN, and Changer-MiT-b1, perform worse under the O-P strategy than under MTKD, and in some cases, even worse than their original models.
This mainly occurs because inaccurate CAR estimation by the original model leads to the selection of an inappropriate teacher model, thereby degrading performance.
More qualitative and quantitative analysis of the performance degradation caused by O-P are provided in Section~\ref{Limitations of O-P}.
This finding highlights the potential drawbacks of the O-P strategy in certain scenarios, but it also demonstrates that our MTKD method can enhance student models' capabilities beyond those of the teacher models trained on partitioned datasets.

\begin{table}[!t]
\centering
\caption{Comparison of Detection Results on Change and No-Change Classes\label{table:Metrics on Different Classes}}
\renewcommand{\arraystretch}{1.25}
\begin{tabular}{cccccc}
\toprule
Method & Class  & IoU  & Rec  & Prec & Fscore    \\ \hline
\cellcolor[HTML]{EFEFEF} & \cellcolor[HTML]{EFEFEF}unchanged 
& \cellcolor[HTML]{EFEFEF}+0.10 
& \cellcolor[HTML]{EFEFEF}-0.60 
& \cellcolor[HTML]{EFEFEF}{\textbf{ +0.65}} 
& \cellcolor[HTML]{EFEFEF}+0.06 \\
\multirow{-2}{*}{\cellcolor[HTML]{EFEFEF}\begin{tabular}[c]{@{}c@{}}SNUNet\end{tabular}} & \cellcolor[HTML]{EFEFEF}changed 
& \cellcolor[HTML]{EFEFEF}{\textbf{+4.21}} 
& \cellcolor[HTML]{EFEFEF}{\textbf{+7.38}} 
& \cellcolor[HTML]{EFEFEF}-0.86 
& \cellcolor[HTML]{EFEFEF}{\textbf{+4.54}} \\ \hline
& unchanged & +0.07 & +0.12 & {\textbf{ -0.01}} & +0.05  \\
\multirow{-2}{*}{\begin{tabular}[c]{@{}c@{}}ChangeFormer\\ (MiT-b1)\end{tabular}} 
& changed   & {\textbf{ +1.68}} & {\textbf{+1.35}} & -0.11   & {\textbf{+1.86}} \\ \hline
\cellcolor[HTML]{EFEFEF}& \cellcolor[HTML]{EFEFEF}unchanged 
& \cellcolor[HTML]{EFEFEF}+0.30 
& \cellcolor[HTML]{EFEFEF}+0.29  
& \cellcolor[HTML]{EFEFEF}+0.08 
& \cellcolor[HTML]{EFEFEF}+0.20 \\
\multirow{-2}{*}{\cellcolor[HTML]{EFEFEF}TinyCD} & \cellcolor[HTML]{EFEFEF}changed   
& \cellcolor[HTML]{EFEFEF}{\textbf{+2.72}} 
& \cellcolor[HTML]{EFEFEF}{\textbf{+4.13}} 
& \cellcolor[HTML]{EFEFEF}{\textbf{+0.16}} 
& \cellcolor[HTML]{EFEFEF}{\textbf{+2.85}} \\ \hline
& unchanged & +0.02 & -0.03 & -0.04 & -0.06 \\
\multirow{-2}{*}{CGNet}  & changed   & {\textbf{+0.88}} & {\textbf{ +0.03}} & {\textbf{+2.04}} & {\textbf{+0.59}} \\ \hline
\cellcolor[HTML]{EFEFEF}& \cellcolor[HTML]{EFEFEF}unchanged 
& \cellcolor[HTML]{EFEFEF}+0.12 
& \cellcolor[HTML]{EFEFEF}+0.23 
& \cellcolor[HTML]{EFEFEF}{\textbf{-0.09}} 
& \cellcolor[HTML]{EFEFEF}+0.07 \\
\multirow{-2}{*}{\cellcolor[HTML]{EFEFEF}\begin{tabular}[c]{@{}c@{}}BAN\end{tabular}} & \cellcolor[HTML]{EFEFEF}changed 
& \cellcolor[HTML]{EFEFEF}{\textbf{+0.70}} 
& \cellcolor[HTML]{EFEFEF}{\textbf{+1.21}} 
& \cellcolor[HTML]{EFEFEF}-1.46 
& \cellcolor[HTML]{EFEFEF}{\textbf{+0.82}} \\ \hline
& unchanged & +0.23 & -0.19 & {\textbf{ +0.45}} & +0.20 \\
\multirow{-2}{*}{TTP} 
& changed & {\textbf{+3.36}} & {\textbf{+5.69}} & -3.99 & {\textbf{+3.39}} \\ 
\bottomrule
\end{tabular}
\end{table}

\subsubsection{Visual Comparison}  
In remote sensing CD, four main challenges are typically encountered: false alarms, missed detections, internal consistency, and boundary completeness. 
To visually assess the effectiveness of the O-P and MTKD strategies in addressing these challenges, we present the visual results of several algorithms in Fig. \ref{fig:visual-v2}, where red indicates missed detections and blue indicates false alarms. 
It can be observed that missed detections, particularly for small object changes, occur most frequently across the original algorithms.
Certain algorithms (e.g., Changer-ResNet-18, CGNet, and TTP) exhibit false alarms as well.
Quantitative analysis reveals that both strategies effectively reduce false alarm and missed detection rates while improving internal consistency and boundary precision. 

\begin{figure*}[!t]
    \centering
    \subfloat[FC-EF]{\includegraphics[width=0.95\linewidth]{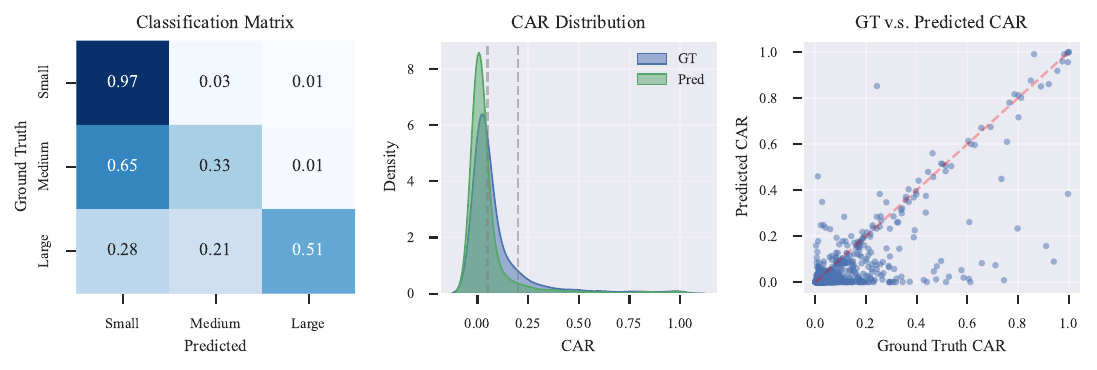}}
    \label{fig:O-P-analysis-FC-EF}
    \\
    \subfloat[STANet]{\includegraphics[width=0.95\linewidth]{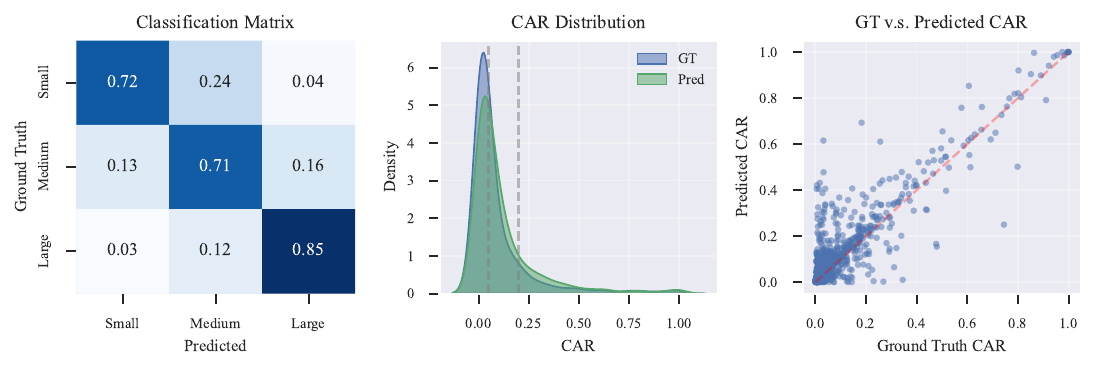}}
    \label{fig:O-P-analysis-STANet}
    \\
    \subfloat[TTP]{\includegraphics[width=0.95\linewidth]{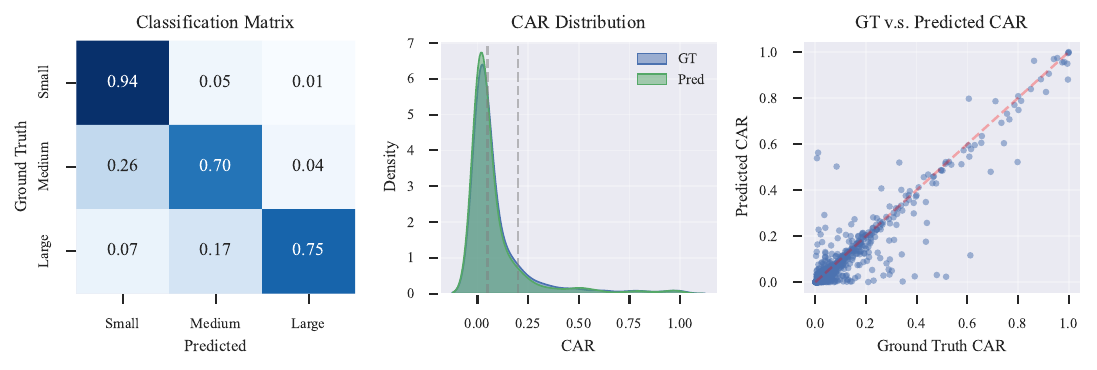}}
    \label{fig:O-P-analysis-TTP}
    \caption{Comparison between estimated CAR from the original model and ground truth CAR values.}
    \label{fig:O-P-analysis}
\end{figure*}

\subsubsection{Results on Different CAR Partitions}  
Fig. \ref{fig:car_per_partition} summarizes the mIoU performance of all benchmark methods across different CAR partitions on the JL1-CD test set. 
The images are sorted by CAR in ascending order and divided into five equal partitions. 
Each bar in the figure represents the lower and upper bounds of CAR for each partition, while the line graph indicates the mIoU across the different partitions. 
The figure indicates that O-P and MTKD yield more notable improvements in partitions with low CAR values, while their impact on high-CAR partitions is generally modest and may even be negative in a few cases (e.g., ChangeFormer-MiT-b1 and TTP).
Taking HANet (CNN), ChangeFormer-MiT-b1 (Transformer), and TTP (FM) as representatives, in the first partition of the test set (CAR range 0.00\% to 0.88\%), O-P improves mIoU by 8.15\%, 3.21\%, and 6.55\%, respectively, while MTKD improves it by 4.42\%, 3.36\%, and 2.85\%. 
This trend is desirable: original models already perform well on large-scale changes (with mIoU often exceeding 70\% in the highest CAR partition), but struggle with small-scale changes (where mIoU frequently falls below 20\% in the lowest CAR partition). 
These results show that by partitioning small CAR regions in the O-P strategy and transferring the knowledge from the corresponding teacher model to the student model in MTKD, our methods achieve substantial improvements in the accuracy of detecting tiny changes.

\subsubsection{Results on Foreground and Background} 
We further select some algorithms and compare their detection performance on change and no-change regions.
The experimental results are summarized in Table \ref{table:Metrics on Different Classes}. 
Taking the original models as baselines, we find that MTKD yields consistently larger IoU and Fscore gains on change areas compared to no-change areas, indicating that the MTKD framework is more sensitive to the detection of change areas.
Interestingly, we observe that MTKD tends to improve recall more than precision. For instance, the recall of the change region in the MTKD-enhanced TTP model improves
by 5.69\%, while precision drops by 3.99\%. Nevertheless, the final F1-score still increases by 3.39\%. This suggests that MTKD effectively reduces the model’s miss rate, and
although it may increase false alarms for some algorithms, the overall performance gains are evident.
More experimental results on the JL1-CD dataset are provided in Appendix~\ref{section: Additional Experiments on the JL1-CD Dataset}.

\begin{table}[!t]
\centering
\caption{Impact of Different Numbers of Teacher Models\label{table:Metrics_with_two_teachers}}
\renewcommand{\arraystretch}{1.25}
\begin{tabular}{ccccc}
\toprule
Method  & Strategy  & $\mathcal{M}_T$ & mIoU  & mFscore \\ \hline
\cellcolor[HTML]{EFEFEF} & \cellcolor[HTML]{EFEFEF} & \cellcolor[HTML]{EFEFEF}3 
& \cellcolor[HTML]{EFEFEF}75.29 (+0.44) 
& \cellcolor[HTML]{EFEFEF}81.32 (+0.34)  \\
\cellcolor[HTML]{EFEFEF} & \multirow{-2}{*}{\cellcolor[HTML]{EFEFEF}O-P} & \cellcolor[HTML]{EFEFEF}2  
& \cellcolor[HTML]{EFEFEF}\textbf{75.44 (+0.59)} 
& \cellcolor[HTML]{EFEFEF}\textbf{81.51 (+0.53)} \\ \cline{2-5}
\cellcolor[HTML]{EFEFEF} & \cellcolor[HTML]{EFEFEF} & \cellcolor[HTML]{EFEFEF}3 
& \cellcolor[HTML]{EFEFEF}75.35 (+0.50) 
& \cellcolor[HTML]{EFEFEF}81.28 (+0.30) \\ 
\multirow{-4}{*}{\cellcolor[HTML]{EFEFEF}\begin{tabular}[c]{@{}c@{}}Changer\\ (MiT-b0)\end{tabular}} 
& \multirow{-2}{*}{\cellcolor[HTML]{EFEFEF}MTKD} & \cellcolor[HTML]{EFEFEF}2 
& \cellcolor[HTML]{EFEFEF}\textbf{75.72 (+0.87)} 
& \cellcolor[HTML]{EFEFEF}\textbf{81.66 (+0.68)} \\ \hline
& & 3 & 72.95 (-0.42) & 79.12 (-0.53) \\
& \multirow{-2}{*}{O-P}  & 2 & \textbf{73.56 (+0.19)} & \textbf{79.92 (+0.27)} \\ \cline{2-5}
& & 3 & \textbf{73.82 (+0.45)} & \textbf{79.91 (+0.26)}  \\
\multirow{-4}{*}{CGNet} 
& \multirow{-2}{*}{MTKD} & 2 & 73.78 (+0.41)  & 79.89 (+0.24) \\ \hline
\cellcolor[HTML]{EFEFEF}& \cellcolor[HTML]{EFEFEF}& \cellcolor[HTML]{EFEFEF}3 
& \cellcolor[HTML]{EFEFEF}{ \textbf{76.69 (+1.64)}} 
& \cellcolor[HTML]{EFEFEF}{ \textbf{82.52 (+1.76)}} \\
\cellcolor[HTML]{EFEFEF} & \multirow{-2}{*}{\cellcolor[HTML]{EFEFEF}O-P}  & \cellcolor[HTML]{EFEFEF}2  
& \cellcolor[HTML]{EFEFEF}{ 76.65 (+1.60)} 
& \cellcolor[HTML]{EFEFEF}{ 82.49 (+1.73)} \\ \cline{2-5}
\cellcolor[HTML]{EFEFEF}& \cellcolor[HTML]{EFEFEF}& \cellcolor[HTML]{EFEFEF}3 
& \cellcolor[HTML]{EFEFEF}\textbf{76.85 (+1.80)} 
& \cellcolor[HTML]{EFEFEF}\textbf{82.56 (+1.80)} \\
\multirow{-4}{*}{\cellcolor[HTML]{EFEFEF}TTP} 
& \multirow{-2}{*}{\cellcolor[HTML]{EFEFEF}MTKD} & \cellcolor[HTML]{EFEFEF}2  
& \cellcolor[HTML]{EFEFEF}76.31 (+1.26) 
& \cellcolor[HTML]{EFEFEF}82.22 (+1.46) \\ \bottomrule
\end{tabular}
\end{table}
\begin{table}[!t]
\centering
\caption{Experimental Results on SYSU-CD Test Set\label{table:metrics_sysucd}}
\renewcommand{\arraystretch}{1.25}
\begin{tabular}{cccccc}
\toprule
Method & Strategy & mIoU  & mRec & mPrec & mFscore \\ \hline
\rowcolor[HTML]{EFEFEF}  \cellcolor[HTML]{EFEFEF}                                             & - & 70.13 & 79.20 & 83.41 & 77.80 \\
\rowcolor[HTML]{EFEFEF} \cellcolor[HTML]{EFEFEF}                                              & O-P & 70.25 & 79.28 & \textbf{83.88} & 77.81 \\
\rowcolor[HTML]{EFEFEF} \multirow{-3}{*}{\cellcolor[HTML]{EFEFEF}SNUNet}                      & MTKD & \textbf{70.61} & \textbf{79.85} & 83.15 & \textbf{78.29} \\ \hline
& - & 71.19 & 79.29 & 87.43 & 78.41 \\
& O-P & 71.76 & 79.84 & 86.62 & 78.80 \\
\multirow{-3}{*}{\begin{tabular}[c]{@{}c@{}}ChangeFormer\\ (MiT-b0)\end{tabular}} 
& MTKD & \textbf{72.16} & \textbf{80.63} & \textbf{86.63} & \textbf{79.44} \\ \hline
\rowcolor[HTML]{EFEFEF} \cellcolor[HTML]{EFEFEF}                                              & - & 76.09 & \textbf{84.59} & 87.08 & 82.72 \\
\rowcolor[HTML]{EFEFEF} \cellcolor[HTML]{EFEFEF}                                              & O-P & 75.97 & 84.54 & 86.20 & 82.66 \\
\rowcolor[HTML]{EFEFEF} \multirow{-3}{*}{\cellcolor[HTML]{EFEFEF}TTP}                         & MTKD & \textbf{76.11} & 83.93 & \textbf{87.77} & \textbf{82.77} \\ \bottomrule
\end{tabular}
\end{table}
\begin{figure*}[!t]
    \centering
    \includegraphics[width=0.95\textwidth]{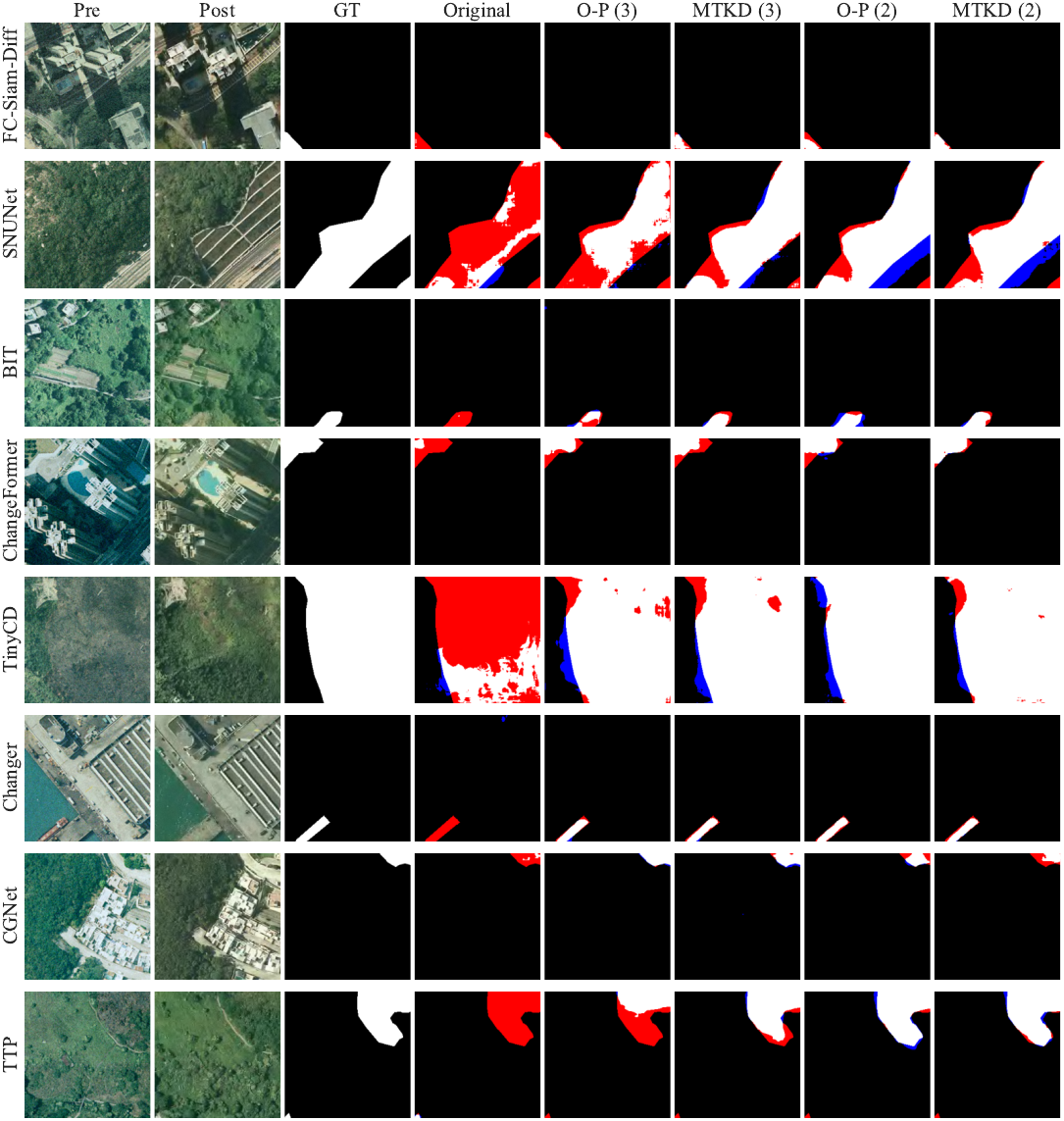}
    \caption{Visual comparison on the SYSU-CD dataset. For each row, the images from left to right are: pre-event image, post-event image, ground truth, output of the original model, output of the O-P strategy with three teachers, output of the MTKD framework with three teachers, output of the O-P strategy with two teachers, and output of the MTKD framework with two teachers. Red indicates missed detections, while blue denotes false alarms. }
    \label{fig:visual-sysucd-more}
\end{figure*}
\subsection{Limitations of O-P} \label{Limitations of O-P}

Whether in Table~\ref{table:metrics} or Fig.~\ref{fig:car_per_partition}, we observe that some models perform worse with the O-P strategy even compared to their original versions. 
This can be attributed to partitioning errors caused by wrong CAR estimation, which may lead to suboptimal model selection during the second-stage inference. 
A deeper analysis of failure scenarios under the O-P strategy is provided.
Fig.~\ref{fig:O-P-analysis} presents the classification matrix, CAR distribution comparison, and scatter plot between the coarse-predicted CAR of the original models and the GT CAR under the O-P strategy for FC-EF, STANet, and TTP.  
For FC-EF, the scatter plot reveals a large number of points where high-GT-CAR samples are mispredicted to have almost zero CAR, which indicates a severe issue of missed detections. This observation is further corroborated by the confusion matrix, where a substantial portion of ``medium'' and ``large'' regions are incorrectly classified as ``small''. 
As a result, the predicted CAR distribution is heavily skewed, with the number of predicted ``small'' samples far exceeding that of the ground truth.  
In contrast, STANet exhibits the opposite trend: its original model tends to generate false alarms: approximately one-quarter of ``small'' samples are misclassified as ``medium'', and more ``medium'' samples are misclassified as ``large'' than as ``small''.  
Such misclassification of CAR regions by the original models leads to suboptimal performance under the O-P strategy, as reflected by the drop in mIoU of $7.03\%$ and $2.20\%$ for FC-EF and STANet, respectively.

On the other hand, the TTP algorithm, already exhibiting strong baseline performance, shows highly accurate CAR partitioning in its original model: the predicted CAR distribution closely matches that of the ground truth. 
Consequently, the O-P strategy provides a notable performance boost, improving mIoU and mFscore by $1.64\%$ and $1.76\%$, respectively.  

These analyses suggest that the effectiveness of the O-P strategy is highly dependent on the quality of the original model's CAR predictions: its benefits are only realized when the baseline is sufficiently strong, while a poor baseline renders it less meaningful. 
Therefore, as a transitional approach, O-P has inherent limitations, whereas the MTKD method, which does not rely on coarse CAR estimation, is more robust and generally preferable.

\subsection{Robustness of MTKD}  
To verify the robustness of the MTKD framework, we conduct the following two sets of experiments:  

\subsubsection{Impact of Teacher Configurations in MTKD}  
In the previous subsection, we have demonstrated that the MTKD framework based on three teacher models can effectively improve the performance of various CD models. 
However, the 3-teacher configuration requires training five versions of the model for a single CD algorithm: $\mathcal{M}_O$, $\mathcal{M}_{T_L}$, $\mathcal{M}_{T_M}$, $\mathcal{M}_{T_S}$, and the student model $\mathcal{M}_S$. 
To alleviate the resource consumption associated with training these models and testing under the O-P strategy, we further explore the performance of the MTKD framework with only two teacher models.
We set a threshold of $th = 0.10$ to divide the dataset into small and large partitions and train the models $\mathcal{M}_{T_S}$ and $\mathcal{M}_{T_L}$ on these partitions, respectively. 
Subsequently, the student model $\mathcal{M}_S$ is trained using the MTKD framework. 
Except for the reduced number of teacher models, all other experimental settings remain unchanged.  
Using the original models as baselines, Table~\ref{table:Metrics_with_two_teachers} compares the metrics of models trained with the O-P strategy and the MTKD framework under different numbers of teacher models. 
The results indicate that two teacher models can still achieve notable performance improvements over the baseline. 
For Changer-MiT-b0, the two-teacher model setup yields even greater improvements compared to the three-teacher setup. 
For CGNet, the O-P strategy with two teachers achieves higher mIoU and mFscore, although the MTKD performance is lower in this case. 
For TTP, the performance of O-P and MTKD under the three teacher model still remains the strongest.
These experiments confirm that the O-P strategy and MTKD framework remain effective with fewer teachers. 

\subsubsection{Performance of MTKD on SYSU-CD} 
We further evaluate the effectiveness of the MTKD framework on the SYSU-CD dataset.
CNN-based SNUNet, Transformer-based ChangeFormer-MiT-b0, and FM-based TTP models are selected.
As summarized in Table~\ref{table:metrics_sysucd}, after MTKD optimization, all the models achieve higher mIoU and mFscore values, with the most notable improvement on ChangeFormer-MiT-b0 (mIoU increased by 0.97\% and mFscore increased by 1.03\%). 
However, the O-P strategy does not yield better results on TTP compared to its original version.  
Overall, MTKD demonstrates substantial effectiveness on the SYSU-CD dataset, further validating its robustness.  
More numerical results on the SYSU-CD dataset are provided in Appendix~\ref{section: Additional Experiments on the SYSU-CD Dataset}.

Visual comparisons are shown in Fig.~\ref{fig:visual-sysucd-more}. Missed detections remain the most common issue for the original models on this dataset. Both the O-P and MTKD strategies effectively mitigate these issues, with MTKD consistently achieving better results.
Nonetheless, limitations remain. For example, while MTKD reduces missed detections in CGNet, some areas remain undetected. Similarly, in TinyCD, although both O-P and MTKD reduce missed detections, a few false positives and holes persist.
These observations highlight promising directions for future research to further improve CD performance.

\section{Conclusion} \label{F}
In this work, we present JL1-CD, a new benchmark dataset for CD, which substantially augments existing datasets by providing sub-meter resolution, a diverse array of change types, and a large dataset scale. 
The MTKD framework is further proposed, which consistently improves the performance of mainstream CD algorithms with various architectures and parameter scales, without incurring additional computational cost during inference.

The proposed methods still present certain limitations. Firstly, the O-P approach heavily relies on the performance of the original model and significantly increases resource consumption during inference. 
MTKD addresses these issues, but a potential limitation of MTKD is that the number of teacher models and the CAR thresholds are determined empirically. Additionally, the resource consumption during the three-stage training process is also considerable.

Future work includes adaptive selection of the number of teacher models and partition thresholds based on different scenarios to further enhance the flexibility and performance of the MTKD framework. Furthermore, optimizing the training process to achieve a balance between accuracy and training efficiency is another avenue for improvement.

\begin{figure*}[!t]
    \centering
    \includegraphics[width=0.98\textwidth]{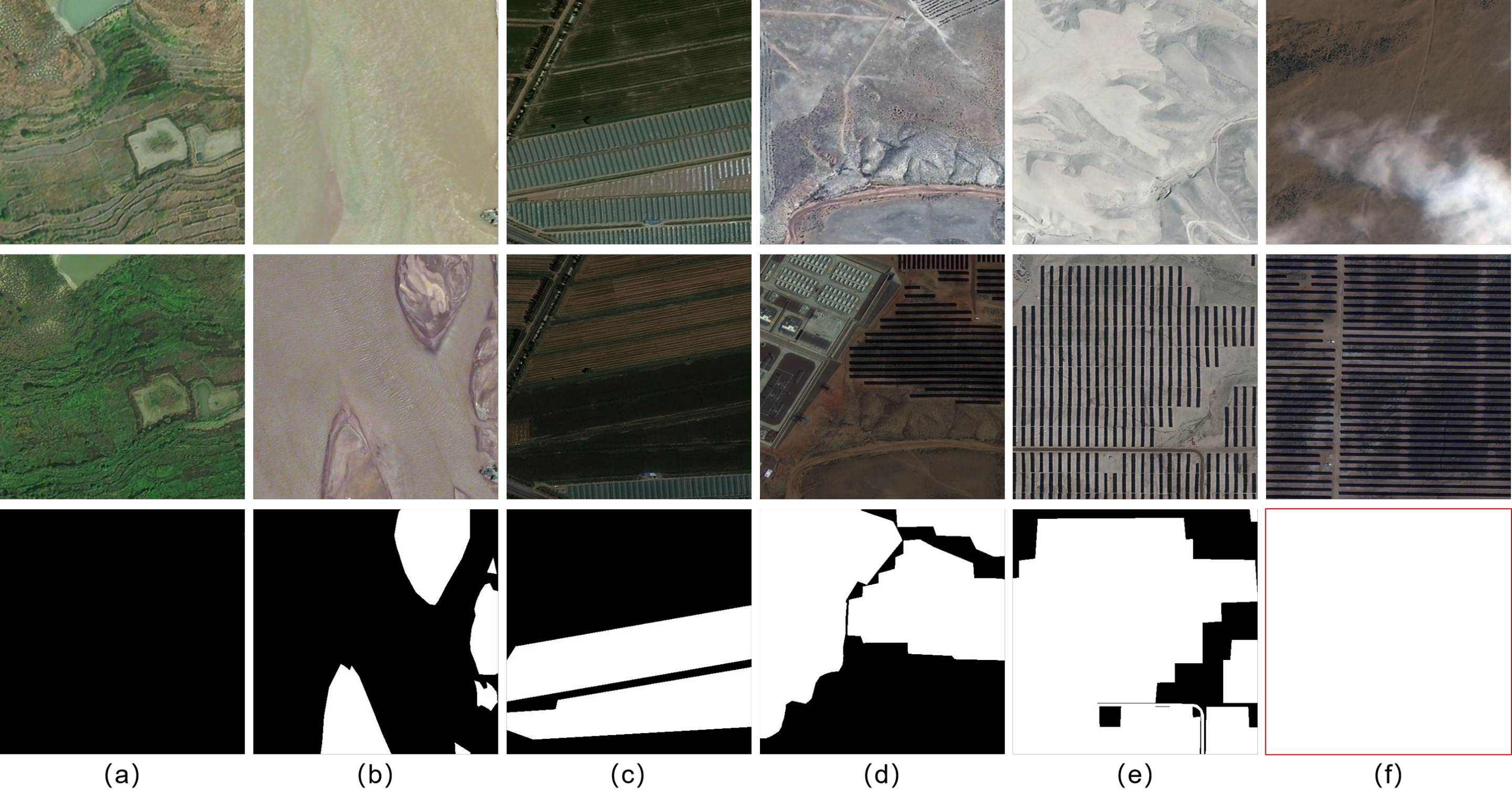}
    \caption{Sample images with different change area ratios (CAR). Each column represents a specific CAR scenario: (a) 0.00\%; (b) 19.98\%; (c) 39.93\%; (d) 59.96\%; (e) 80.25\%; and (f) 100.00\%.}
    \label{fig:dataset}
\end{figure*}

\begin{figure*}[!t]
    \centering
    \subfloat[JL1-CD]{\includegraphics[width=0.48\linewidth]{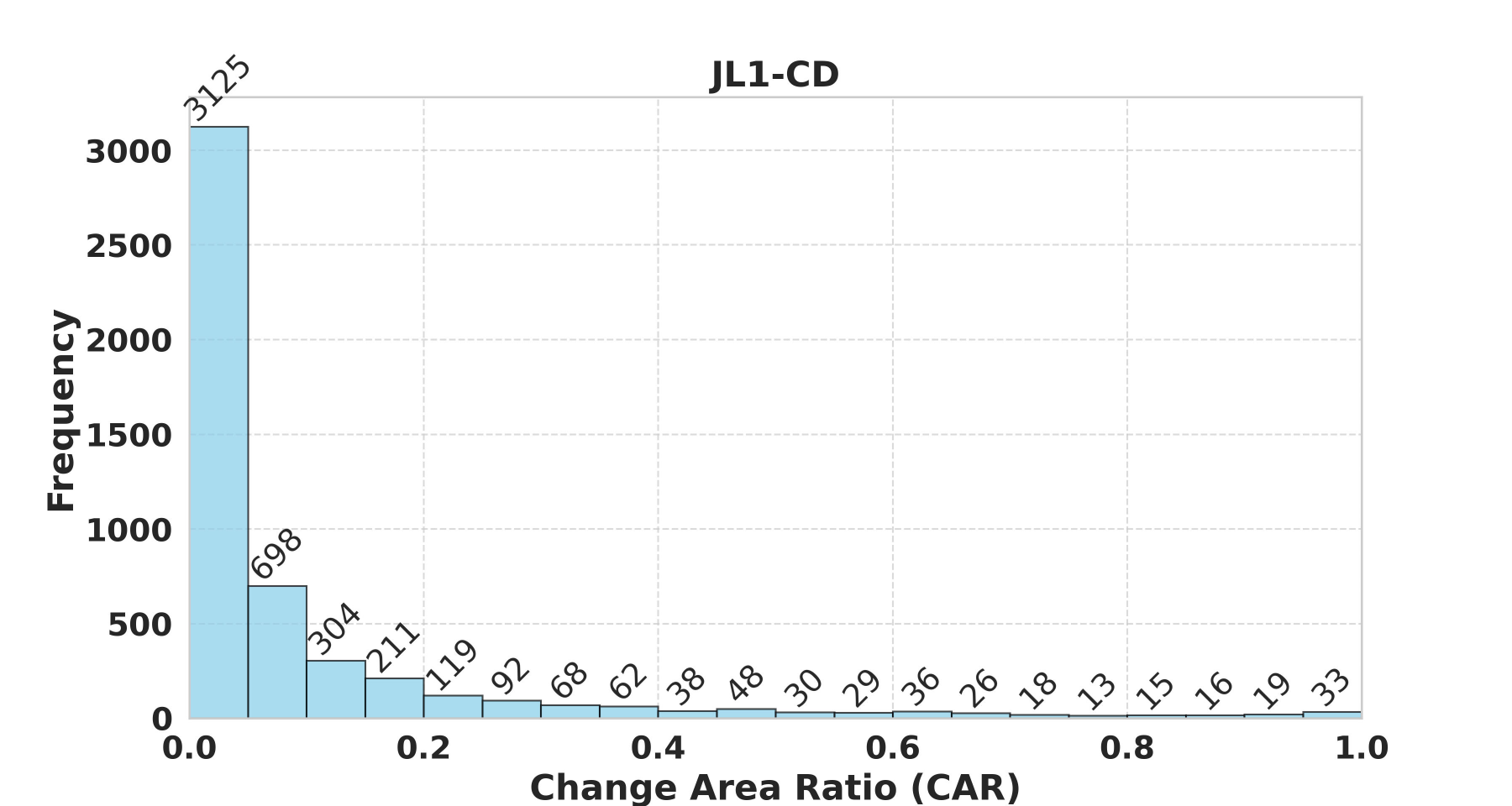}}
    \label{fig:CAR-JL1-CD}
    \subfloat[SYSU-CD]{\includegraphics[width=0.48\linewidth]{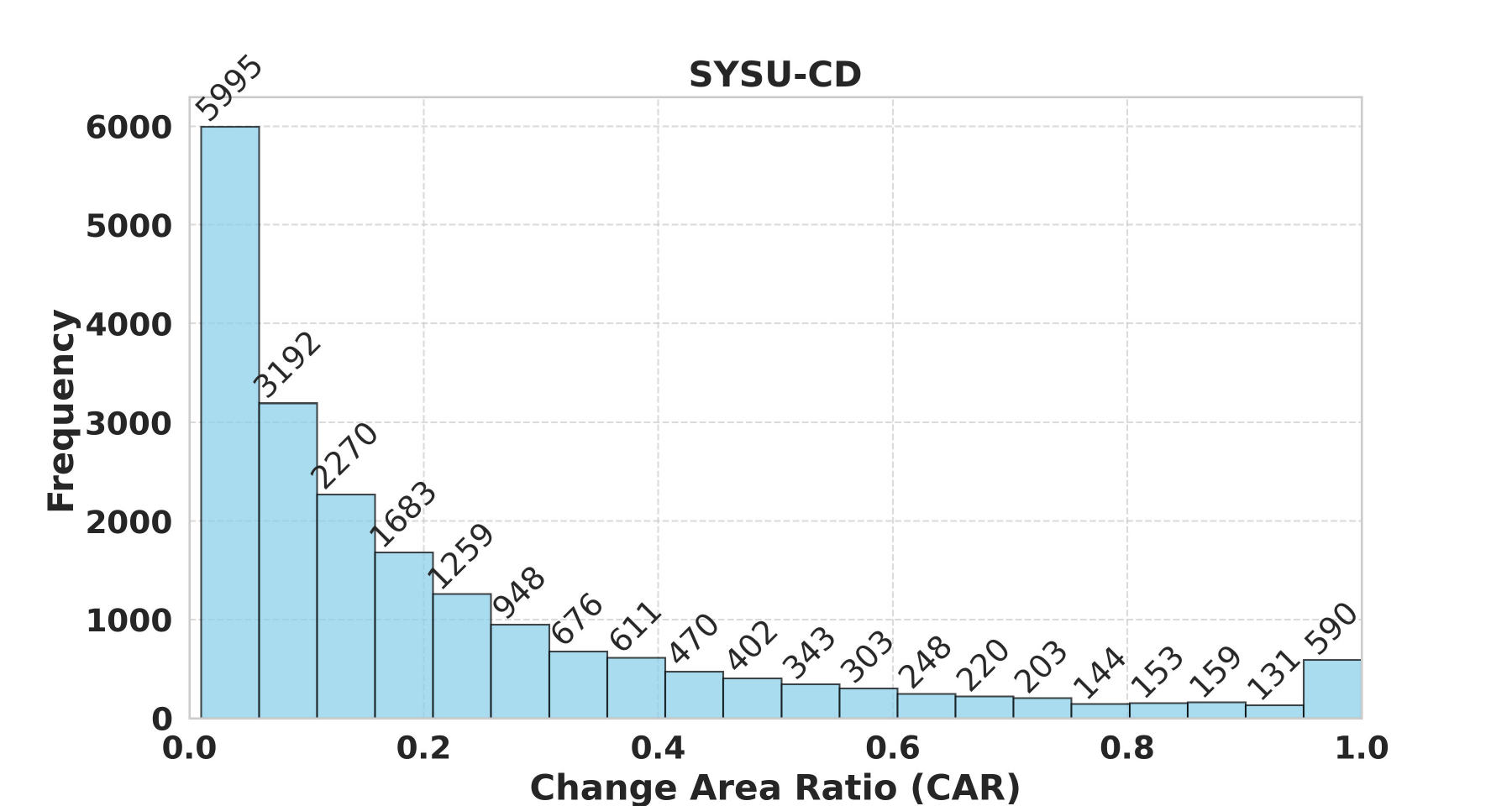}}
    \label{fig:CAR-SYSU-CD}
    \caption{CAR distribution of the JL1-CD and SYSU-CD datasets.}
    \label{fig:CAR-JL1-CD-SYSU-CD}
\end{figure*}

\appendix
\subsection{CAR Distribution} \label{section: car distribution}
Fig.~\ref{fig:dataset} presents sample images from the JL1-CD dataset with varying values of the Change Area Ratio (CAR). We further provide a quantitative analysis of the CAR distributions for the JL1-CD and SYSU-CD datasets in Fig.~\ref{fig:CAR-JL1-CD-SYSU-CD}. 
The CAR spans from 0 to 100\% in these datasets, which poses significant challenges for model learning, and it serves as the primary motivation behind our proposed O-P strategy and MTKD framework.

\subsection{Additional Experiments on the JL1-CD Dataset} \label{section: Additional Experiments on the JL1-CD Dataset}
We summarize the quantitative results of 21 models on both change and no-change regions in Table~\ref{table:metrics-more} to provide a comprehensive evaluation of the JL1-CD dataset as a new benchmark and to further validate the effectiveness and generality of the MTKD framework.
All models exhibit improvements in IoU and Fscore for change regions after applying MTKD. 

\begin{table*}[!t]
\centering
\caption{Experimental Results on JL1-CD Test Set (to Better Illustrate the Impact of the O-P Strategy and MTKD Framework on the Detection Performance of Change and No-Change Classes, Only the Relative Improvements in Evaluation Metrics Compared to the Original Baseline Model are Reported) 
\label{table:metrics-more}}
\renewcommand{\arraystretch}{1.05}
\setlength{\tabcolsep}{3.8pt}
\begin{tabular}{ccccccccccccccc}
\toprule
Method & Strategy & Class & IoU & Rec & Prec & Fscore & & 
Method & Strategy & Class & IoU & Rec & Prec & Fscore \\ \hline
\cellcolor[HTML]{EFEFEF}  & \cellcolor[HTML]{EFEFEF} & \cellcolor[HTML]{EFEFEF}unchanged & \cellcolor[HTML]{EFEFEF}92.85 & \cellcolor[HTML]{EFEFEF}98.45  & \cellcolor[HTML]{EFEFEF}94.01 & \cellcolor[HTML]{EFEFEF}95.56 &  & \cellcolor[HTML]{EFEFEF} & \cellcolor[HTML]{EFEFEF} & \cellcolor[HTML]{EFEFEF}unchanged & \cellcolor[HTML]{EFEFEF}93.95 & \cellcolor[HTML]{EFEFEF}97.89  & \cellcolor[HTML]{EFEFEF}95.70  & \cellcolor[HTML]{EFEFEF}96.30 \\
\cellcolor[HTML]{EFEFEF} & \multirow{-2}{*}{\cellcolor[HTML]{EFEFEF}-} & \cellcolor[HTML]{EFEFEF}changed   & \cellcolor[HTML]{EFEFEF}21.30  & \cellcolor[HTML]{EFEFEF}25.36  & \cellcolor[HTML]{EFEFEF}78.79  & \cellcolor[HTML]{EFEFEF}27.00  &  & \cellcolor[HTML]{EFEFEF}                                                & \multirow{-2}{*}{\cellcolor[HTML]{EFEFEF}-}    & \cellcolor[HTML]{EFEFEF}changed & \cellcolor[HTML]{EFEFEF}33.64 & \cellcolor[HTML]{EFEFEF}41.19  & \cellcolor[HTML]{EFEFEF}73.84  & \cellcolor[HTML]{EFEFEF}42.08 \\ \cline{2-7} \cline{10-15} 
\cellcolor[HTML]{EFEFEF} & \cellcolor[HTML]{EFEFEF} & \cellcolor[HTML]{EFEFEF}unchanged & \cellcolor[HTML]{EFEFEF}-0.10  & \cellcolor[HTML]{EFEFEF}+0.48  & \cellcolor[HTML]{EFEFEF}-0.45  & \cellcolor[HTML]{EFEFEF}+0.08  &  & \cellcolor[HTML]{EFEFEF} & \cellcolor[HTML]{EFEFEF} & \cellcolor[HTML]{EFEFEF}unchanged & \cellcolor[HTML]{EFEFEF}+0.05 & \cellcolor[HTML]{EFEFEF}+0.38  & \cellcolor[HTML]{EFEFEF}-0.25  & \cellcolor[HTML]{EFEFEF}+0.13 \\
\cellcolor[HTML]{EFEFEF} & \multirow{-2}{*}{\cellcolor[HTML]{EFEFEF}O-P}  & \cellcolor[HTML]{EFEFEF}changed   & \cellcolor[HTML]{EFEFEF}-13.95 & \cellcolor[HTML]{EFEFEF}-17.02 & \cellcolor[HTML]{EFEFEF}+18.20 & \cellcolor[HTML]{EFEFEF}-18.64 &  & \cellcolor[HTML]{EFEFEF} & \multirow{-2}{*}{\cellcolor[HTML]{EFEFEF}O-P}  & \cellcolor[HTML]{EFEFEF}changed   & \cellcolor[HTML]{EFEFEF}-5.78 & \cellcolor[HTML]{EFEFEF}-10.02 & \cellcolor[HTML]{EFEFEF}+12.77 & \cellcolor[HTML]{EFEFEF}-7.59 \\ \cline{2-7} \cline{10-15} 
\cellcolor[HTML]{EFEFEF}  & \cellcolor[HTML]{EFEFEF} & \cellcolor[HTML]{EFEFEF}unchanged & \cellcolor[HTML]{EFEFEF}+0.47  & \cellcolor[HTML]{EFEFEF}-0.95  & \cellcolor[HTML]{EFEFEF}+1.40  & \cellcolor[HTML]{EFEFEF}+0.39  &  & \cellcolor[HTML]{EFEFEF} & \cellcolor[HTML]{EFEFEF} & \cellcolor[HTML]{EFEFEF}unchanged & \cellcolor[HTML]{EFEFEF}-0.03 & \cellcolor[HTML]{EFEFEF}-0.42 & \cellcolor[HTML]{EFEFEF}+0.34  & \cellcolor[HTML]{EFEFEF}+0.06 \\
\multirow{-6}{*}{\cellcolor[HTML]{EFEFEF}FC-EF} & \multirow{-2}{*}{\cellcolor[HTML]{EFEFEF}MTKD} & \cellcolor[HTML]{EFEFEF}changed   & \cellcolor[HTML]{EFEFEF}+9.12  & \cellcolor[HTML]{EFEFEF}+14.04 & \cellcolor[HTML]{EFEFEF}-10.40 & \cellcolor[HTML]{EFEFEF}+11.72 &  & \multirow{-6}{*}{\cellcolor[HTML]{EFEFEF}FC-EF} & \multirow{-2}{*}{\cellcolor[HTML]{EFEFEF}MTKD} & \cellcolor[HTML]{EFEFEF}changed & \cellcolor[HTML]{EFEFEF}+1.05 & \cellcolor[HTML]{EFEFEF}+2.49  & \cellcolor[HTML]{EFEFEF}-2.36  & \cellcolor[HTML]{EFEFEF}+1.32 \\
& & unchanged & 93.74 & 98.71 & 94.77 & 96.17 &  &  &  & unchanged & 92.27 & 94.22 & 97.64 & 95.41 \\
& \multirow{-2}{*}{-} & changed & 28.86 & 33.34 & 78.13 & 36.51 & & & \multirow{-2}{*}{-} & changed & 41.26 & 69.21 & 51.82 & 52.63 \\ \cline{2-7} \cline{10-15} 
& & unchanged & +0.26 & -0.44 & +0.68 & +0.26 & & & & unchanged & -0.06 & +0.79 & -0.86 & -0.04 \\
& \multirow{-2}{*}{O-P} & changed & -1.00 & -2.17 & +8.48 & -2.02 & &  & \multirow{-2}{*}{O-P}              & changed & -4.36 & -7.28 & -4.35 & -5.49 \\ \cline{2-7} \cline{10-15} 
& & unchanged & +0.15 & -0.28 & +0.36 & +0.10 & & & & unchanged & +0.65 & +0.74 & -0.12 & +0.33 \\
\multirow{-6}{*}{FC-Siam-Diff} & \multirow{-2}{*}{MTKD} & changed & +4.76 & +7.36 & -5.35 & +6.21 &  & \multirow{-6}{*}{STANet} & \multirow{-2}{*}{MTKD} & changed & +1.66 & -0.03 & +3.13 & +1.83 \\
\cellcolor[HTML]{EFEFEF} & \cellcolor[HTML]{EFEFEF} & \cellcolor[HTML]{EFEFEF}unchanged & \cellcolor[HTML]{EFEFEF}94.59  & \cellcolor[HTML]{EFEFEF}97.34  & \cellcolor[HTML]{EFEFEF}96.93  & \cellcolor[HTML]{EFEFEF}96.76  &  & \cellcolor[HTML]{EFEFEF} & \cellcolor[HTML]{EFEFEF} & \cellcolor[HTML]{EFEFEF}unchanged & \cellcolor[HTML]{EFEFEF}94.59 & \cellcolor[HTML]{EFEFEF}98.12  & \cellcolor[HTML]{EFEFEF}96.21  & \cellcolor[HTML]{EFEFEF}96.76 \\
\cellcolor[HTML]{EFEFEF} & \multirow{-2}{*}{\cellcolor[HTML]{EFEFEF}-} & \cellcolor[HTML]{EFEFEF}changed   & \cellcolor[HTML]{EFEFEF}47.90 & \cellcolor[HTML]{EFEFEF}60.48  & \cellcolor[HTML]{EFEFEF}72.14  & \cellcolor[HTML]{EFEFEF}57.89 & & \cellcolor[HTML]{EFEFEF} & \multirow{-2}{*}{\cellcolor[HTML]{EFEFEF}-}    & \cellcolor[HTML]{EFEFEF}changed & \cellcolor[HTML]{EFEFEF}43.34 & \cellcolor[HTML]{EFEFEF}51.63  & \cellcolor[HTML]{EFEFEF}73.92 & \cellcolor[HTML]{EFEFEF}53.75 \\ \cline{2-7} \cline{10-15} 
\cellcolor[HTML]{EFEFEF} & \cellcolor[HTML]{EFEFEF} & \cellcolor[HTML]{EFEFEF}unchanged & \cellcolor[HTML]{EFEFEF}-0.17 & \cellcolor[HTML]{EFEFEF}-0.23  & \cellcolor[HTML]{EFEFEF}+0.12  & \cellcolor[HTML]{EFEFEF}-0.06 & & \cellcolor[HTML]{EFEFEF} & \cellcolor[HTML]{EFEFEF} & \cellcolor[HTML]{EFEFEF}unchanged & \cellcolor[HTML]{EFEFEF}-0.06 & \cellcolor[HTML]{EFEFEF}-0.18  & \cellcolor[HTML]{EFEFEF}+0.03  & \cellcolor[HTML]{EFEFEF}-0.03 \\
\cellcolor[HTML]{EFEFEF}  & \multirow{-2}{*}{\cellcolor[HTML]{EFEFEF}O-P} & \cellcolor[HTML]{EFEFEF}changed & \cellcolor[HTML]{EFEFEF}-0.19 & \cellcolor[HTML]{EFEFEF}-0.85  & \cellcolor[HTML]{EFEFEF}-0.63  & \cellcolor[HTML]{EFEFEF}-0.17  &  & \cellcolor[HTML]{EFEFEF}  & \multirow{-2}{*}{\cellcolor[HTML]{EFEFEF}O-P}  & \cellcolor[HTML]{EFEFEF}changed   & \cellcolor[HTML]{EFEFEF}+4.91 & \cellcolor[HTML]{EFEFEF}+7.63  & \cellcolor[HTML]{EFEFEF}-3.44  & \cellcolor[HTML]{EFEFEF}+5.49 \\ \cline{2-7} \cline{10-15} 
\cellcolor[HTML]{EFEFEF} & \cellcolor[HTML]{EFEFEF} & \cellcolor[HTML]{EFEFEF}unchanged & \cellcolor[HTML]{EFEFEF}+0.24  & \cellcolor[HTML]{EFEFEF}+0.29  & \cellcolor[HTML]{EFEFEF}+0.04  & \cellcolor[HTML]{EFEFEF}+0.12  &  & \cellcolor[HTML]{EFEFEF}  & \cellcolor[HTML]{EFEFEF} & \cellcolor[HTML]{EFEFEF}unchanged & \cellcolor[HTML]{EFEFEF}+0.10 & \cellcolor[HTML]{EFEFEF}-0.60  & \cellcolor[HTML]{EFEFEF}+0.65  & \cellcolor[HTML]{EFEFEF}+0.06 \\
\multirow{-6}{*}{\cellcolor[HTML]{EFEFEF}IFN} & \multirow{-2}{*}{\cellcolor[HTML]{EFEFEF}MTKD} & \cellcolor[HTML]{EFEFEF}changed   & \cellcolor[HTML]{EFEFEF}+2.71  & \cellcolor[HTML]{EFEFEF}+2.44  & \cellcolor[HTML]{EFEFEF}+0.22  & \cellcolor[HTML]{EFEFEF}+2.82  &  & \multirow{-6}{*}{\cellcolor[HTML]{EFEFEF}SNUNet} & \multirow{-2}{*}{\cellcolor[HTML]{EFEFEF}MTKD} & \cellcolor[HTML]{EFEFEF}changed   & \cellcolor[HTML]{EFEFEF}+4.21 & \cellcolor[HTML]{EFEFEF}+7.38  & \cellcolor[HTML]{EFEFEF}-0.86  & \cellcolor[HTML]{EFEFEF}+4.54 \\
& & unchanged & 94.14 & 97.34 & 96.47 & 96.50 & & & & unchanged & 94.91 & 98.04 & 96.62 & 97.00 \\
& \multirow{-2}{*}{-} & changed & 40.31 & 51.61 & 70.95 & 50.25 &  & & \multirow{-2}{*}{-} & changed & 44.04 & 53.12 & 72.31 & 54.15 \\ \cline{2-7} \cline{10-15} 
& & unchanged & +0.13 & +0.03 & +0.14 & +0.05 & & & & unchanged & -0.48 & -0.62 & +0.02 & -0.35 \\
& \multirow{-2}{*}{O-P} & changed & +4.23 & +3.59 & +0.49 & +4.75 & & & \multirow{-2}{*}{O-P} & changed & -0.73 & -1.05 & +0.86 & -1.08 \\ \cline{2-7} \cline{10-15} 
& & unchanged & +0.42 & +0.66 & -0.10 & +0.22 & & & & unchanged & +0.13 & -0.06 & +0.19 & +0.09 \\
\multirow{-6}{*}{BIT} & \multirow{-2}{*}{MTKD} & changed & +2.86 & +1.37 & +2.09 & +2.79 &  & \multirow{-6}{*}{\begin{tabular}[c]{@{}c@{}}ChangeStar\\ (FarSeg)\end{tabular}} & \multirow{-2}{*}{MTKD} & changed & +0.65 & +0.51 & +2.44 & +0.82 \\
\cellcolor[HTML]{EFEFEF} & \cellcolor[HTML]{EFEFEF} & \cellcolor[HTML]{EFEFEF}unchanged & \cellcolor[HTML]{EFEFEF}94.36  & \cellcolor[HTML]{EFEFEF}98.44  & \cellcolor[HTML]{EFEFEF}95.67  & \cellcolor[HTML]{EFEFEF}96.57  &  & \cellcolor[HTML]{EFEFEF} & \cellcolor[HTML]{EFEFEF} & \cellcolor[HTML]{EFEFEF}unchanged 
& \cellcolor[HTML]{EFEFEF}95.28 
& \cellcolor[HTML]{EFEFEF}98.26  
& \cellcolor[HTML]{EFEFEF}96.83  
& \cellcolor[HTML]{EFEFEF}97.19 \\
\cellcolor[HTML]{EFEFEF} & \multirow{-2}{*}{\cellcolor[HTML]{EFEFEF}-} & \cellcolor[HTML]{EFEFEF}changed   & \cellcolor[HTML]{EFEFEF}35.35  & \cellcolor[HTML]{EFEFEF}39.93  & \cellcolor[HTML]{EFEFEF}80.84  & \cellcolor[HTML]{EFEFEF}43.81  &  & \cellcolor[HTML]{EFEFEF} & \multirow{-2}{*}{\cellcolor[HTML]{EFEFEF}-}    & \cellcolor[HTML]{EFEFEF}changed   
& \cellcolor[HTML]{EFEFEF}51.74 
& \cellcolor[HTML]{EFEFEF}62.66  
& \cellcolor[HTML]{EFEFEF}75.82  
& \cellcolor[HTML]{EFEFEF}62.21 \\ \cline{2-7} \cline{10-15} 
\cellcolor[HTML]{EFEFEF} & \cellcolor[HTML]{EFEFEF} & \cellcolor[HTML]{EFEFEF}unchanged & \cellcolor[HTML]{EFEFEF}+0.15  & \cellcolor[HTML]{EFEFEF}-0.10  & \cellcolor[HTML]{EFEFEF}+0.20  & \cellcolor[HTML]{EFEFEF}+0.21  &  & \cellcolor[HTML]{EFEFEF} & \cellcolor[HTML]{EFEFEF} & \cellcolor[HTML]{EFEFEF}unchanged 
& \cellcolor[HTML]{EFEFEF}+0.13 
& \cellcolor[HTML]{EFEFEF}-0.14  
& \cellcolor[HTML]{EFEFEF}+0.24  
& \cellcolor[HTML]{EFEFEF}+0.13 \\
\cellcolor[HTML]{EFEFEF} & \multirow{-2}{*}{\cellcolor[HTML]{EFEFEF}O-P}  & \cellcolor[HTML]{EFEFEF}changed   & \cellcolor[HTML]{EFEFEF}-0.49  & \cellcolor[HTML]{EFEFEF}-0.16  & \cellcolor[HTML]{EFEFEF}-2.26  & \cellcolor[HTML]{EFEFEF}-0.42  &  & \cellcolor[HTML]{EFEFEF} & \multirow{-2}{*}{\cellcolor[HTML]{EFEFEF}O-P}  & \cellcolor[HTML]{EFEFEF}changed   
& \cellcolor[HTML]{EFEFEF}-1.99 
& \cellcolor[HTML]{EFEFEF}-2.46  
& \cellcolor[HTML]{EFEFEF}-0.22  
& \cellcolor[HTML]{EFEFEF}-1.95 \\ \cline{2-7} \cline{10-15} 
\cellcolor[HTML]{EFEFEF} & \cellcolor[HTML]{EFEFEF} & \cellcolor[HTML]{EFEFEF}unchanged & \cellcolor[HTML]{EFEFEF}-0.08  & \cellcolor[HTML]{EFEFEF}-0.16  
& \cellcolor[HTML]{EFEFEF}+0.08  & \cellcolor[HTML]{EFEFEF}-0.03  &  & \cellcolor[HTML]{EFEFEF} & \cellcolor[HTML]{EFEFEF} & \cellcolor[HTML]{EFEFEF}unchanged 
& \cellcolor[HTML]{EFEFEF}+0.26 
& \cellcolor[HTML]{EFEFEF}-0.18  
& \cellcolor[HTML]{EFEFEF}+0.47  
& \cellcolor[HTML]{EFEFEF}+0.15  \\
\multirow{-6}{*}{\cellcolor[HTML]{EFEFEF}\begin{tabular}[c]{@{}c@{}}ChangeStar\\ (UPerNet)\end{tabular}} & \multirow{-2}{*}{\cellcolor[HTML]{EFEFEF}MTKD} & \cellcolor[HTML]{EFEFEF}changed   & \cellcolor[HTML]{EFEFEF}+0.57  & \cellcolor[HTML]{EFEFEF}+2.31  & \cellcolor[HTML]{EFEFEF}-1.20  & \cellcolor[HTML]{EFEFEF}+0.81  &  & \multirow{-6}{*}{\cellcolor[HTML]{EFEFEF}\begin{tabular}[c]{@{}c@{}}ChangeFormer\\ (MiT-b0)\end{tabular}} & \multirow{-2}{*}{\cellcolor[HTML]{EFEFEF}MTKD} & \cellcolor[HTML]{EFEFEF}changed   
& \cellcolor[HTML]{EFEFEF}+0.28 
& \cellcolor[HTML]{EFEFEF}-0.60  
& \cellcolor[HTML]{EFEFEF}+0.69  
& \cellcolor[HTML]{EFEFEF}+0.40 \\
& & unchanged & 95.40 & 98.08 & 97.12 & 97.32 & & & & unchanged & 94.23 & 97.10 & 96.54 & 96.43 \\
& \multirow{-2}{*}{-} & changed & 50.70 & 61.31 & 76.78 & 61.13 & & & \multirow{-2}{*}{-} & changed & 47.85 & 60.44 & 69.55 & 59.05 \\ \cline{2-7} \cline{10-15} 
& & unchanged & +0.04 & +0.17 & -0.11 & +0.00 &  & & & unchanged & +0.30 & +0.39 & +0.19 & +0.29 \\
 & \multirow{-2}{*}{O-P} & changed & +0.77 & -1.17 & +1.12 & +0.38 &  & & \multirow{-2}{*}{O-P} & changed & +2.06 & +1.92 & +0.69 & +1.76 \\ \cline{2-7} \cline{10-15} 
 & & unchanged & +0.07 & +0.12 & -0.01 & +0.05 &  & & & unchanged & +0.30 & +0.29 & +0.08 & +0.20 \\
\multirow{-6}{*}{\begin{tabular}[c]{@{}c@{}}ChangeFormer\\ (MiT-b1)\end{tabular}} & \multirow{-2}{*}{MTKD} & changed & +1.68 & +1.35 & -0.11 & +1.86 &  & \multirow{-6}{*}{TinyCD} & \multirow{-2}{*}{MTKD} & changed & +2.72 & +4.13 & +0.16 & +2.85 \\
\cellcolor[HTML]{EFEFEF} & \cellcolor[HTML]{EFEFEF} & \cellcolor[HTML]{EFEFEF}unchanged & \cellcolor[HTML]{EFEFEF}93.53  & \cellcolor[HTML]{EFEFEF}97.48  & \cellcolor[HTML]{EFEFEF}95.51  & \cellcolor[HTML]{EFEFEF}96.04  &  & \cellcolor[HTML]{EFEFEF} & \cellcolor[HTML]{EFEFEF} & \cellcolor[HTML]{EFEFEF}unchanged & \cellcolor[HTML]{EFEFEF}95.37 & \cellcolor[HTML]{EFEFEF}97.55  & \cellcolor[HTML]{EFEFEF}97.57  & \cellcolor[HTML]{EFEFEF}97.20 \\
\cellcolor[HTML]{EFEFEF} & \multirow{-2}{*}{\cellcolor[HTML]{EFEFEF}-} & \cellcolor[HTML]{EFEFEF}changed   & \cellcolor[HTML]{EFEFEF}33.76  & \cellcolor[HTML]{EFEFEF}42.06  & \cellcolor[HTML]{EFEFEF}71.35  & \cellcolor[HTML]{EFEFEF}42.74  &  & \cellcolor[HTML]{EFEFEF} & \multirow{-2}{*}{\cellcolor[HTML]{EFEFEF}-}    & \cellcolor[HTML]{EFEFEF}changed   & \cellcolor[HTML]{EFEFEF}54.32 & \cellcolor[HTML]{EFEFEF}66.13  & \cellcolor[HTML]{EFEFEF}74.61  & \cellcolor[HTML]{EFEFEF}64.76 \\ \cline{2-7} \cline{10-15} 
\cellcolor[HTML]{EFEFEF} & \cellcolor[HTML]{EFEFEF} & \cellcolor[HTML]{EFEFEF}unchanged & \cellcolor[HTML]{EFEFEF}+0.67  & \cellcolor[HTML]{EFEFEF}+0.21  & \cellcolor[HTML]{EFEFEF}+0.60  & \cellcolor[HTML]{EFEFEF}+0.51  &  & \cellcolor[HTML]{EFEFEF} & \cellcolor[HTML]{EFEFEF} & \cellcolor[HTML]{EFEFEF}unchanged & \cellcolor[HTML]{EFEFEF}+0.04 & \cellcolor[HTML]{EFEFEF}+0.31  & \cellcolor[HTML]{EFEFEF}-0.33  & \cellcolor[HTML]{EFEFEF}+0.10 \\
\cellcolor[HTML]{EFEFEF} & \multirow{-2}{*}{\cellcolor[HTML]{EFEFEF}O-P}  & \cellcolor[HTML]{EFEFEF}changed   & \cellcolor[HTML]{EFEFEF}+10.14 & \cellcolor[HTML]{EFEFEF}+13.32 & \cellcolor[HTML]{EFEFEF}-1.37  & \cellcolor[HTML]{EFEFEF}+12.02 &  & \cellcolor[HTML]{EFEFEF} & \multirow{-2}{*}{\cellcolor[HTML]{EFEFEF}O-P}  & \cellcolor[HTML]{EFEFEF}changed   & \cellcolor[HTML]{EFEFEF}+0.84 & \cellcolor[HTML]{EFEFEF}-1.19  & \cellcolor[HTML]{EFEFEF}+2.27  & \cellcolor[HTML]{EFEFEF}+0.59 \\ \cline{2-7} \cline{10-15} 
\cellcolor[HTML]{EFEFEF} & \cellcolor[HTML]{EFEFEF} & \cellcolor[HTML]{EFEFEF}unchanged & \cellcolor[HTML]{EFEFEF}+0.78  & \cellcolor[HTML]{EFEFEF}+0.28  & \cellcolor[HTML]{EFEFEF}+0.61  & \cellcolor[HTML]{EFEFEF}+0.53  &  & \cellcolor[HTML]{EFEFEF} & \cellcolor[HTML]{EFEFEF} & \cellcolor[HTML]{EFEFEF}unchanged & \cellcolor[HTML]{EFEFEF}+0.21 & \cellcolor[HTML]{EFEFEF}+0.32  & \cellcolor[HTML]{EFEFEF}-0.14  & \cellcolor[HTML]{EFEFEF}+0.19 \\
\multirow{-6}{*}{\cellcolor[HTML]{EFEFEF}HANet} & \multirow{-2}{*}{\cellcolor[HTML]{EFEFEF}MTKD} & \cellcolor[HTML]{EFEFEF}changed   & \cellcolor[HTML]{EFEFEF}+7.27  & \cellcolor[HTML]{EFEFEF}+8.96  & \cellcolor[HTML]{EFEFEF}+1.29  & \cellcolor[HTML]{EFEFEF}+8.53  &  & \multirow{-6}{*}{\cellcolor[HTML]{EFEFEF}\begin{tabular}[c]{@{}c@{}}Changer\\ (MiT-b0)\end{tabular}}      & \multirow{-2}{*}{\cellcolor[HTML]{EFEFEF}MTKD} & \cellcolor[HTML]{EFEFEF}changed   & \cellcolor[HTML]{EFEFEF}+0.80 & \cellcolor[HTML]{EFEFEF}-0.48  & \cellcolor[HTML]{EFEFEF}+2.32  & \cellcolor[HTML]{EFEFEF}+0.41 \\
 & & unchanged & 95.87 & 98.17 & 97.51 & 97.62 &  & & & unchanged & 94.07 & 97.58 & 96.13 & 96.39 \\
 & \multirow{-2}{*}{-} & changed & 56.01 & 65.81 & 77.96 & 66.24 &  & & \multirow{-2}{*}{-} & changed & 42.66 & 52.71 & 70.73 & 52.68 \\ \cline{2-7} \cline{10-15} 
 & & unchanged & -0.44 & -0.10 & -0.35 & -0.30 &  & & & unchanged & +0.58 & -0.16 & +0.65 & +0.38 \\
 & \multirow{-2}{*}{O-P} & changed & -0.61 & -0.54 & -0.86 & -0.69 &  & & \multirow{-2}{*}{O-P} & changed & +4.21 & +4.71 & +0.21 & +4.56 \\ \cline{2-7} \cline{10-15} 
 & & unchanged & +0.02 & +0.09 & -0.04 & -0.01 &  & & & unchanged & +0.11 & -0.33 & +0.40 & +0.07 \\
\multirow{-6}{*}{\begin{tabular}[c]{@{}c@{}}Changer\\ (MiT-b1)\end{tabular}} & \multirow{-2}{*}{MTKD} & changed & +0.41 & +1.63 & -1.47 & +0.42 &  & \multirow{-6}{*}{\begin{tabular}[c]{@{}c@{}}Changer\\ (ResNet-18)\end{tabular}} & \multirow{-2}{*}{MTKD} & changed & +2.07 & +4.57 & -4.26 & +2.57 \\
\cellcolor[HTML]{EFEFEF} & \cellcolor[HTML]{EFEFEF} & \cellcolor[HTML]{EFEFEF}unchanged & \cellcolor[HTML]{EFEFEF}92.90  & \cellcolor[HTML]{EFEFEF}96.42  & \cellcolor[HTML]{EFEFEF}95.90  & \cellcolor[HTML]{EFEFEF}95.66  &  & \cellcolor[HTML]{EFEFEF} & \cellcolor[HTML]{EFEFEF} & \cellcolor[HTML]{EFEFEF}unchanged 
& \cellcolor[HTML]{EFEFEF}93.52
& \cellcolor[HTML]{EFEFEF}96.65
& \cellcolor[HTML]{EFEFEF}96.52
& \cellcolor[HTML]{EFEFEF}96.05 \\
\cellcolor[HTML]{EFEFEF} & \multirow{-2}{*}{\cellcolor[HTML]{EFEFEF}-} & \cellcolor[HTML]{EFEFEF}changed   & \cellcolor[HTML]{EFEFEF}31.72  & \cellcolor[HTML]{EFEFEF}42.03  & \cellcolor[HTML]{EFEFEF}65.92  & \cellcolor[HTML]{EFEFEF}40.00  &  & \cellcolor[HTML]{EFEFEF} & \multirow{-2}{*}{\cellcolor[HTML]{EFEFEF}-}    & \cellcolor[HTML]{EFEFEF}changed 
& \cellcolor[HTML]{EFEFEF}35.63
& \cellcolor[HTML]{EFEFEF}45.66
& \cellcolor[HTML]{EFEFEF}69.71
& \cellcolor[HTML]{EFEFEF}44.21 \\ \cline{2-7} \cline{10-15} 
\cellcolor[HTML]{EFEFEF} & \cellcolor[HTML]{EFEFEF} & \cellcolor[HTML]{EFEFEF}unchanged & \cellcolor[HTML]{EFEFEF}+1.24  & \cellcolor[HTML]{EFEFEF}+0.68  & \cellcolor[HTML]{EFEFEF}+0.73  & \cellcolor[HTML]{EFEFEF}+0.73  &  & \cellcolor[HTML]{EFEFEF} & \cellcolor[HTML]{EFEFEF} & \cellcolor[HTML]{EFEFEF}unchanged 
& \cellcolor[HTML]{EFEFEF}+0.54 
& \cellcolor[HTML]{EFEFEF}+0.31  
& \cellcolor[HTML]{EFEFEF}+0.23  
& \cellcolor[HTML]{EFEFEF}+0.33 \\
\cellcolor[HTML]{EFEFEF} & \multirow{-2}{*}{\cellcolor[HTML]{EFEFEF}O-P}  & \cellcolor[HTML]{EFEFEF}changed   & \cellcolor[HTML]{EFEFEF}+17.73 & \cellcolor[HTML]{EFEFEF}+20.40 & \cellcolor[HTML]{EFEFEF}+3.75  & \cellcolor[HTML]{EFEFEF}+20.07 &  & \cellcolor[HTML]{EFEFEF} & \multirow{-2}{*}{\cellcolor[HTML]{EFEFEF}O-P}  & \cellcolor[HTML]{EFEFEF}changed   
& \cellcolor[HTML]{EFEFEF}+10.69 
& \cellcolor[HTML]{EFEFEF}+12.24  
& \cellcolor[HTML]{EFEFEF}+1.51  
& \cellcolor[HTML]{EFEFEF}+11.73 \\ \cline{2-7} \cline{10-15} 
\cellcolor[HTML]{EFEFEF} & \cellcolor[HTML]{EFEFEF} & \cellcolor[HTML]{EFEFEF}unchanged & \cellcolor[HTML]{EFEFEF}+0.40  & \cellcolor[HTML]{EFEFEF}+0.48  & \cellcolor[HTML]{EFEFEF}-0.06  & \cellcolor[HTML]{EFEFEF}+0.27  &  & \cellcolor[HTML]{EFEFEF} & \cellcolor[HTML]{EFEFEF} & \cellcolor[HTML]{EFEFEF}unchanged 
& \cellcolor[HTML]{EFEFEF}+0.65 
& \cellcolor[HTML]{EFEFEF}+0.87  
& \cellcolor[HTML]{EFEFEF}-0.32  
& \cellcolor[HTML]{EFEFEF}+0.45 \\
\multirow{-6}{*}{\cellcolor[HTML]{EFEFEF}\begin{tabular}[c]{@{}c@{}}Changer\\ (ResNeSt-50)\end{tabular}} & \multirow{-2}{*}{\cellcolor[HTML]{EFEFEF}MTKD} & \cellcolor[HTML]{EFEFEF}changed   & \cellcolor[HTML]{EFEFEF}+0.89  & \cellcolor[HTML]{EFEFEF}+0.36  & \cellcolor[HTML]{EFEFEF}+1.77  & \cellcolor[HTML]{EFEFEF}+1.12  &  & \multirow{-6}{*}{\cellcolor[HTML]{EFEFEF}LightCDNet} & \multirow{-2}{*}{\cellcolor[HTML]{EFEFEF}MTKD} & \cellcolor[HTML]{EFEFEF}changed   
& \cellcolor[HTML]{EFEFEF}+2.19 
& \cellcolor[HTML]{EFEFEF}+1.70  
& \cellcolor[HTML]{EFEFEF}+1.81  
& \cellcolor[HTML]{EFEFEF}+2.25 \\
 & & unchanged & 95.34 & 98.23 & 96.93 & 97.30 &  & & & unchanged & 95.57 & 98.14 & 97.26 & 97.40 \\
 & \multirow{-2}{*}{-} & changed & 51.40 & 62.40 & 73.74 & 62.00 &  & & \multirow{-2}{*}{-} & changed & 51.51 & 60.94 & 78.52 & 61.55 \\ \cline{2-7} \cline{10-15} 
 & & unchanged & -0.38 & -0.69 & +0.24 & -0.27 &  & & & unchanged & +0.01 & +0.39 & -0.46 & +0.00 \\
 & \multirow{-2}{*}{O-P} & changed & -0.47 & -0.53 & +0.08 & -0.78 &  & & \multirow{-2}{*}{O-P} & changed & +0.14 & -1.12 & +0.87 & -0.04 \\ \cline{2-7} \cline{10-15} 
 & & unchanged & +0.02 & -0.03 & -0.04 & -0.06 &  & & & unchanged & +0.12 & +0.23 & -0.09 & +0.07 \\
\multirow{-6}{*}{CGNet} & \multirow{-2}{*}{MTKD} & changed & +0.88 & +0.03 & +2.04  & +0.59 & & \multirow{-6}{*}{BAN} & \multirow{-2}{*}{MTKD} & changed & +0.70 & +1.21 & -1.46 & +0.82 \\
\cellcolor[HTML]{EFEFEF} & \cellcolor[HTML]{EFEFEF} & \cellcolor[HTML]{EFEFEF}unchanged & \cellcolor[HTML]{EFEFEF}95.68  & \cellcolor[HTML]{EFEFEF}98.46  & \cellcolor[HTML]{EFEFEF}97.03  & \cellcolor[HTML]{EFEFEF}97.41  &  & & & & & & & \\
\cellcolor[HTML]{EFEFEF} & \multirow{-2}{*}{\cellcolor[HTML]{EFEFEF}-}    & \cellcolor[HTML]{EFEFEF}changed   & \cellcolor[HTML]{EFEFEF}54.43  & \cellcolor[HTML]{EFEFEF}62.02  & \cellcolor[HTML]{EFEFEF}82.62  & \cellcolor[HTML]{EFEFEF}64.12  &  &  & & & & & & \\ \cline{2-7}
\cellcolor[HTML]{EFEFEF} & \cellcolor[HTML]{EFEFEF} & \cellcolor[HTML]{EFEFEF}unchanged & \cellcolor[HTML]{EFEFEF}-0.14  & \cellcolor[HTML]{EFEFEF}-0.56  & \cellcolor[HTML]{EFEFEF}+0.39  & \cellcolor[HTML]{EFEFEF}-0.06  &  & & & & & & & \\
\cellcolor[HTML]{EFEFEF} & \multirow{-2}{*}{\cellcolor[HTML]{EFEFEF}O-P}  & \cellcolor[HTML]{EFEFEF}changed   & \cellcolor[HTML]{EFEFEF}+3.42  & \cellcolor[HTML]{EFEFEF}+7.03  & \cellcolor[HTML]{EFEFEF}-5.50  & \cellcolor[HTML]{EFEFEF}+3.58  &  & & & & & & & \\ \cline{2-7}
\cellcolor[HTML]{EFEFEF} & \cellcolor[HTML]{EFEFEF} & \cellcolor[HTML]{EFEFEF}unchanged & \cellcolor[HTML]{EFEFEF}+0.23  & \cellcolor[HTML]{EFEFEF}-0.19  & \cellcolor[HTML]{EFEFEF}+0.45  & \cellcolor[HTML]{EFEFEF}+0.20  &  & & & & & & & \\
\multirow{-6}{*}{\cellcolor[HTML]{EFEFEF}TTP} & \multirow{-2}{*}{\cellcolor[HTML]{EFEFEF}MTKD} & \cellcolor[HTML]{EFEFEF}changed & \cellcolor[HTML]{EFEFEF}+3.36  & \cellcolor[HTML]{EFEFEF}+5.69  & \cellcolor[HTML]{EFEFEF}-3.99  & \cellcolor[HTML]{EFEFEF}+3.39  &  & & & & & & & \\ \bottomrule
\end{tabular}
\end{table*}

\begin{table*}[!t]
\centering
\caption{Experimental Results on SYSU-CD Test Set (Best and Second-Best Results are in Bold and Underlined, Respectively)\label{table:metrics_sysucd_full}}
\renewcommand{\arraystretch}{1.25}
\begin{tabular}{cccccccccccc}
\toprule
Method & Strategy & $\mathcal{M}_T$ & mIoU & mRec & mPrec & mFscore & Class & IoU & Rec & Prec & Fscore \\ \hline
\multirow{6}{*}{FC-Siam-Diff} & \multirow{2}{*}{-} & \multirow{2}{*}{-} & \multirow{2}{*}{62.04} & \multirow{2}{*}{70.78} & \multirow{2}{*}{\textbf{85.80}} & \multirow{2}{*}{69.10} & unchanged & 83.76 & 97.02  & 85.82 & 89.12  \\
& & & & & & & changed & 40.32 & 44.54 & 85.78 & 49.09  \\ \cline{2-12} 
& \multirow{4}{*}{MTKD} & \multirow{2}{*}{3} & \multirow{2}{*}{{\ul 64.04}} & \multirow{2}{*}{{\ul 72.83}} & \multirow{2}{*}{{\ul 85.29}} & \multirow{2}{*}{{\ul 71.13}} 
& unchanged & +0.61 & -0.59  & +1.13 & +0.45  \\
& & & & & & & changed & +3.39 & +4.70 & -2.16 & +3.61  \\ \cline{3-12} 
 & & \multirow{2}{*}{2} & \multirow{2}{*}{\textbf{65.22}} & \multirow{2}{*}{\textbf{75.14}} & \multirow{2}{*}{82.83} & \multirow{2}{*}{\textbf{72.86}} & unchanged & -0.19 & -2.60  & +1.87 & -0.12  \\
 & & & & & & & changed   & +6.56 & +11.33 & -7.82 & +7.63  \\ \hline
\multirow{6}{*}{SNUNet} & \multirow{2}{*}{-}    & \multirow{2}{*}{-} & \multirow{2}{*}{70.13} & \multirow{2}{*}{79.20} & \multirow{2}{*}{{\ul 83.41}} & \multirow{2}{*}{77.80} & unchanged & 85.53 & 93.79  & 90.37 & 90.45  \\
 & & & & & & & changed & 54.73 & 64.62  & 76.45 & 65.15  \\ \cline{2-12} 
 & \multirow{4}{*}{MTKD} & \multirow{2}{*}{3} & \multirow{2}{*}{\textbf{70.61}} & \multirow{2}{*}{\textbf{79.85}} & \multirow{2}{*}{83.15}          & \multirow{2}{*}{\textbf{78.29}} & unchanged & +0.06 & -0.43  & +0.37 & -0.06  \\
 & & & & & & & changed   & +0.90 & +1.72  & -0.89 & +1.04  \\ \cline{3-12} 
 & & \multirow{2}{*}{2} & \multirow{2}{*}{{\ul 70.43}} & \multirow{2}{*}{{\ul 79.46}}    & \multirow{2}{*}{\textbf{83.97}} & \multirow{2}{*}{{\ul 78.08}} & unchanged & -0.05 & -0.28  & +0.41 & -0.22  \\
 & & & & & & & changed   & +0.65 & +0.79  & +0.71 & +0.78  \\ \hline
\multirow{6}{*}{BIT} & \multirow{2}{*}{-} & \multirow{2}{*}{-} & \multirow{2}{*}{67.16} & \multirow{2}{*}{76.03} & \multirow{2}{*}{\textbf{84.27}} & \multirow{2}{*}{74.76} & unchanged & 83.90 & 92.22  & 90.25 & 88.89  \\
 & & & & & & & changed   & 50.42 & 59.85  & 78.30 & 60.62  \\ \cline{2-12} 
 & \multirow{4}{*}{MTKD} & \multirow{2}{*}{3} & \multirow{2}{*}{{\ul 68.80}} & \multirow{2}{*}{{\ul 78.52}} & \multirow{2}{*}{82.43} & \multirow{2}{*}{{\ul 76.56}}    & unchanged & +0.35 & -0.35  & +0.84 & +0.49  \\
 & & & & & & & changed & +2.92 & +5.32  & -4.52 & +3.12  \\ \cline{3-12} 
 & & \multirow{2}{*}{2} & \multirow{2}{*}{\textbf{68.97}} & \multirow{2}{*}{\textbf{78.71}} & \multirow{2}{*}{{\ul 83.10}} & \multirow{2}{*}{\textbf{76.83}} & unchanged & +0.28 & -0.52  & +0.88 & +0.35  \\
 & & & & & & & changed   & +3.34 & +5.87  & -3.23 & 64.41  \\ \hline
\multirow{6}{*}{\begin{tabular}[c]{@{}c@{}}ChangeFormer\\ (MiT-b0)\end{tabular}} & \multirow{2}{*}{-}    & \multirow{2}{*}{-} & \multirow{2}{*}{71.19} & \multirow{2}{*}{79.29} & \multirow{2}{*}{\textbf{87.43}} & \multirow{2}{*}{78.41} & unchanged & 86.33 & 94.32  & 91.03 & 90.82  \\
 & & & & & & & changed   & 56.06 & 64.26  & 83.83 & 66.00  \\ \cline{2-12} 
 & \multirow{4}{*}{MTKD} & \multirow{2}{*}{3} & \multirow{2}{*}{\textbf{72.16}} & \multirow{2}{*}{{\ul 80.63}} & \multirow{2}{*}{{\ul 86.62}} & \multirow{2}{*}{\textbf{79.44}} & unchanged & +0.33 & -0.34  & +0.52 & +0.34  \\
 & & & & & & & changed   & +1.60 & +3.02  & -2.12 & +1.72  \\ \cline{3-12} 
 & & \multirow{2}{*}{2} & \multirow{2}{*}{{\ul 71.95}} & \multirow{2}{*}{\textbf{80.65}} & \multirow{2}{*}{85.61} & \multirow{2}{*}{{\ul 79.28}} & unchanged & +0.36 & +0.02  & +0.27 & +0.38  \\
 & & & & & & & changed & +1.15 & +2.69  & -3.91 & +1.36  \\ \hline
\multirow{6}{*}{TinyCD} & \multirow{2}{*}{-} & \multirow{2}{*}{-} & \multirow{2}{*}{70.68} & \multirow{2}{*}{\textbf{80.28}} & \multirow{2}{*}{83.59} & \multirow{2}{*}{78.42} & unchanged & 85.14 & 93.96  & 89.91 & 89.92  \\
 & & & & & & & changed & 56.22 & 66.59  & 77.27 & 66.92  \\ \cline{2-12} 
 & \multirow{4}{*}{MTKD} & \multirow{2}{*}{3} & \multirow{2}{*}{{\ul 70.94}} & \multirow{2}{*}{80.11} & \multirow{2}{*}{\textbf{84.14}} & \multirow{2}{*}{{\ul 78.59}}    & unchanged & -0.01 & -0.99  & +0.82 & -0.07  \\
 & & & & & & & changed   & +0.53 & +0.66  & +0.28 & +0.40  \\ \cline{3-12} 
 & & \multirow{2}{*}{2} & \multirow{2}{*}{\textbf{70.97}} & \multirow{2}{*}{{\ul 80.22}}    & \multirow{2}{*}{{\ul 83.78}}    & \multirow{2}{*}{\textbf{78.70}} & unchanged & -0.10 & -1.18  & +0.86 & -0.13  \\
 & & & & & & & changed   & +0.69 & +1.07  & -0.48 & +0.69  \\ \hline
\multirow{6}{*}{\begin{tabular}[c]{@{}c@{}}Changer\\ (MiT-b1)\end{tabular}} & \multirow{2}{*}{-}    & \multirow{2}{*}{-} & \multirow{2}{*}{75.49} & \multirow{2}{*}{{\ul 84.25}} & \multirow{2}{*}{86.38} & \multirow{2}{*}{82.38} & unchanged & 87.75 & 93.93  & 92.86 & 91.92  \\
 & & & & & & & changed   & 63.23 & 74.58  & 79.91 & 72.85  \\ \cline{2-12} 
 & \multirow{4}{*}{MTKD} & \multirow{2}{*}{3} & \multirow{2}{*}{{\ul 75.56}} & \multirow{2}{*}{83.97} & \multirow{2}{*}{\textbf{87.33}} & \multirow{2}{*}{{\ul 82.40}}    & unchanged & +0.15 & +0.44  & -0.12 & +0.08  \\
 & & & & & & & changed   & +0.00 & -1.02  & +2.02 & -0.05  \\ \cline{3-12} 
 & & \multirow{2}{*}{2} & \multirow{2}{*}{\textbf{75.96}} & \multirow{2}{*}{\textbf{84.31}} & \multirow{2}{*}{{\ul 87.03}}    & \multirow{2}{*}{\textbf{82.76}} & unchanged & +0.37 & +0.20  & +0.38 & +0.29  \\
 & & & & & & & changed   & +0.56 & -0.09  & +0.92 & +0.45  \\ \hline
\multirow{6}{*}{CGNet} & \multirow{2}{*}{-}    & \multirow{2}{*}{-} & \multirow{2}{*}{71.41} & \multirow{2}{*}{80.32} & \multirow{2}{*}{\textbf{84.72}} & \multirow{2}{*}{78.82} & unchanged & 86.08 & 93.96  & 90.96 & 90.83  \\
 & & & & & & & changed   & 56.75 & 66.67  & 78.49 & 66.81  \\ \cline{2-12} 
 & \multirow{4}{*}{MTKD} & \multirow{2}{*}{3} & \multirow{2}{*}{\textbf{71.87}} & \multirow{2}{*}{\textbf{81.74}} & \multirow{2}{*}{84.05} & \multirow{2}{*}{\textbf{79.42}} & unchanged & -0.49 & -2.18  & +1.57 & -0.42  \\
 & & & & & & & changed   & +1.40 & +5.03  & -2.91 & +1.61  \\ \cline{3-12} 
 & & \multirow{2}{*}{2} & \multirow{2}{*}{{\ul 71.75}} & \multirow{2}{*}{{\ul 80.64}} & \multirow{2}{*}{{\ul 84.69}} & \multirow{2}{*}{{\ul 79.21}}    & unchanged & +0.22 & -0.01  & +0.10 & +0.18  \\
 & & & & & & & changed   & +0.44 & +0.66  & -0.17 & +0.59  \\ \hline
\multirow{6}{*}{TTP} & \multirow{2}{*}{-}    & \multirow{2}{*}{-} & \multirow{2}{*}{76.09} & \multirow{2}{*}{{\ul 84.59}} & \multirow{2}{*}{{\ul 87.08}} & \multirow{2}{*}{82.72} & unchanged & 88.30 & 94.46  & 93.08 & 92.26  \\
 & & & & & & & changed   & 63.88 & 74.71  & 81.08 & 73.17  \\ \cline{2-12} 
 & \multirow{4}{*}{MTKD} & \multirow{2}{*}{3} & \multirow{2}{*}{{\ul 76.11}} & \multirow{2}{*}{83.93} & \multirow{2}{*}{\textbf{87.77}} & \multirow{2}{*}{{\ul 82.77}}    & unchanged & +0.05 & +0.65  & -0.73 & +0.03  \\
 & & & & & & & changed   & +0.00 & -1.96  & +2.11 & +0.07  \\ \cline{3-12} 
 & & \multirow{2}{*}{2} & \multirow{2}{*}{\textbf{76.53}} & \multirow{2}{*}{\textbf{85.54}} & \multirow{2}{*}{86.22} & \multirow{2}{*}{\textbf{83.29}} & unchanged & -0.26 & -0.85  & +0.41 & -0.19  \\
 & & & & & & & changed   & +1.14 & +2.75  & -2.13 & +1.34  \\ \bottomrule
\end{tabular}
\end{table*}

\subsection{Additional Experiments on the SYSU-CD Dataset} \label{section: Additional Experiments on the SYSU-CD Dataset}
The SYSU-CD dataset is a large-scale and challenging dataset with a wide CAR range. Therefore, we select it as the second benchmark to evaluate the robustness of our proposed methods. 
Except for changing the patch size to 256 × 256, all other experimental settings are kept identical to those used for experiments on JL1-CD.

Table~\ref{table:metrics_sysucd_full} presents additional quantitative results under different teacher model configurations, including detection metrics for both the change and no-change classes. As summarized in Table~\ref{table:metrics_sysucd_full}, our MTKD approach achieves consistent and notable improvements. 
After applying MTKD, all models, regardless of whether two or three teacher models are used, exhibit increased mIoU and mFscore values. 
Specifically, the FC-Siam-Diff model trained with two teacher models achieves the highest improvements in mIoU (+3.18\%) and mFscore (+3.76\%). 

Among the evaluated models, SNUNet, ChangeFormer-MiT-b0, and CGNet demonstrate superior performance under the three-teacher setting, whereas FC-Siam-Diff, BiT, TinyCD, Changer-MiT-b1, and TTP perform better with two teachers.
This observation suggests the need for future work on methods that can automatically select the optimal number of teachers depending on the scenario.

In the last four columns of Table~\ref{table:metrics_sysucd_full}, we report detection metrics separately for the change and no-change classes. The absolute values correspond to the original models, while the MTKD framework results are presented as relative improvements.
We observe more substantial gains in detecting change areas. For instance, in the two-teacher setup, FC-Siam-Diff’s IoU and Fscore on the change class increase by 6.56\% and 7.63\%, respectively, whereas performance on the no-change class decreases slightly. 
These results suggest that MTKD enhances sensitivity toward change regions, consistent with our findings on the JL1-CD dataset.

\bibliographystyle{IEEEtran}
\bibliography{refs}

\end{document}